\RequirePackage{fix-cm}
\documentclass[twocolumn,natbib]{svjour3}          
\smartqed  
\usepackage{cite}
\usepackage{graphicx}
\usepackage{amsmath}
\usepackage{amsfonts}
\usepackage{booktabs}       

\usepackage{xcolor}
\usepackage{algorithm2e}
\RestyleAlgo{ruled}
\usepackage{multirow} 
\usepackage{float} 

\usepackage{nicefrac} 
\usepackage{microtype}   
\usepackage{bm}
\usepackage{bbm}
\usepackage{pifont}
\usepackage{subfigure}
\usepackage{wrapfig}
\usepackage{multicol}
\usepackage{amssymb}
\usepackage{pdfpages}
\usepackage{makecell}
\usepackage{xspace}
\usepackage{threeparttable}
\usepackage{soul}
\usepackage{colortbl}
\usepackage{arydshln}

\usepackage[colorlinks, citecolor=blue]{hyperref}
\makeatletter
\def\UrlAlphabet{%
      \do\a\do\b\do\c\do\d\do\e\do\f\do\g\do\h\do\i\do\j%
      \do\k\do\l\do\m\do\n\do\o\do\p\do\q\do\r\do\s\do\t%
      \do\u\do\v\do\w\do\x\do\y\do\z\do\A\do\B\do\C\do\D%
      \do\E\do\F\do\G\do\H\do\I\do\J\do\K\do\L\do\M\do\N%
      \do\O\do\P\do\Q\do\R\do\S\do\T\do\U\do\V\do\W\do\X%
      \do\Y\do\Z}
\def\UrlDigits{\do\1\do\2\do\3\do\4\do\5\do\6\do\7\do\8\do\9\do\0}
\g@addto@macro{\UrlBreaks}{\UrlOrds}
\g@addto@macro{\UrlBreaks}{\UrlAlphabet}
\g@addto@macro{\UrlBreaks}{\UrlDigits}
\makeatother

\makeatletter
\DeclareRobustCommand\onedot{\futurelet\@let@token\@onedot}
\def\@onedot{\ifx\@let@token.\else.\null\fi\xspace}

\def\eg{\emph{e.g}\onedot} 
\def\ie{\emph{i.e}\onedot} 
 
\def\etc{\emph{etc}\onedot} 
 
\def\etal{\emph{et al}\onedot}
\makeatother

\def\ie{{\em i.e.}}
\def\eg{{\em e.g.}}
\def\etal{{\em et al.}}

\usepackage{balance}
\usepackage[misc]{ifsym} 
\usepackage[colorlinks, citecolor=blue]{hyperref}

\journalname{International Journal of Computer Vision}
\begin{document}

\title{Advancing Weakly-Supervised Audio-Visual Video Parsing via Segment-wise Pseudo Labeling\thanks{This work was supported by the National Key R\&D Program of China (NO.2022YFB4500601), the National Natural Science Foundation of China (72188101, 62272144, 62020106007, and U20A20183), the Major Project of Anhui Province (202203a05020011), and the Fundamental Research Funds for the Central Universities. This work is also partially supported by the National Key R\&D Program of China (NO.2022ZD0160100).}
}

\author{Jinxing Zhou$^{1,2}$,
        Dan Guo$^{1,3,4*}$,
Yiran Zhong$^{2}$,
        Meng Wang$^{1,3*}$
}

\authorrunning{Jinxing Zhou et. al} 

\institute{
$^{*}$: Corresponding authors \vspace{2ex}\\ $^{\textrm{\Letter}}$:\\ Jinxing Zhou \\
\hspace{2ex}{zhoujxhfut@gmail.com}
\vspace{1ex}\\
Dan Guo\\
guodan@hfut.edu.cn
\vspace{1ex}\\
Yiran Zhong\\
zhongyiran@gmail.com\vspace{1ex}\\
Meng Wang\\
eric.mengwang@gmail.com 
\vspace{2ex}
\\
$^{1}$: 
Hefei University of Technology, Hefei, China \\ 
$^{2}$: Shanghai AI Laboratory, Shanghai, China\\
$^{3}$: 
Hefei Comprehensive National Science Center, Hefei, China\\
$^{4}$: Anhui Zhonghuitong Technology Co., Ltd., Hefei, China
}

\date{Received: date / Accepted: date}

\maketitle

\begin{abstract}
 {The Audio-Visual Video Parsing task aims to identify and temporally localize the events that occur in either or both the audio and visual streams of audible videos.} It often performs in a weakly-supervised manner, where only video event labels are provided, \ie, the modalities and the timestamps of the labels are unknown. Due to the lack of densely annotated labels, recent work attempts to leverage pseudo labels to enrich the supervision. A commonly used strategy is to generate pseudo labels by categorizing the known video event labels for each modality. However, the labels are still confined to the video level, and the temporal boundaries of events remain unlabeled. In this paper, we propose a new pseudo label generation strategy that can explicitly assign labels to each video segment by utilizing prior knowledge learned from the open world. Specifically, we exploit the large-scale pretrained models, namely CLIP and CLAP, to estimate the events in each video segment and generate segment-level visual and audio pseudo labels, respectively. We then propose a new loss function to exploit these pseudo labels by taking into account their category-richness and segment-richness. A label denoising strategy is also adopted to further improve the visual pseudo labels by flipping them whenever abnormally large forward losses occur. We perform extensive experiments on the LLP dataset and demonstrate the effectiveness of each proposed design and we achieve state-of-the-art video parsing performance on all types of event parsing, \ie, audio event, visual event, and audio-visual event.
Furthermore, our experiments verify that the high-quality segment-level pseudo labels provided by our method can be flexibly combined with other audio-visual video parsing backbones and consistently improve their performances.
We also examine the proposed pseudo label generation strategy on a relevant weakly-supervised audio-visual event localization task and the experimental results again verify the benefits and generalization of our method.
\end{abstract}

\begin{figure*}[t]
  \centering
\includegraphics[width=\textwidth]
{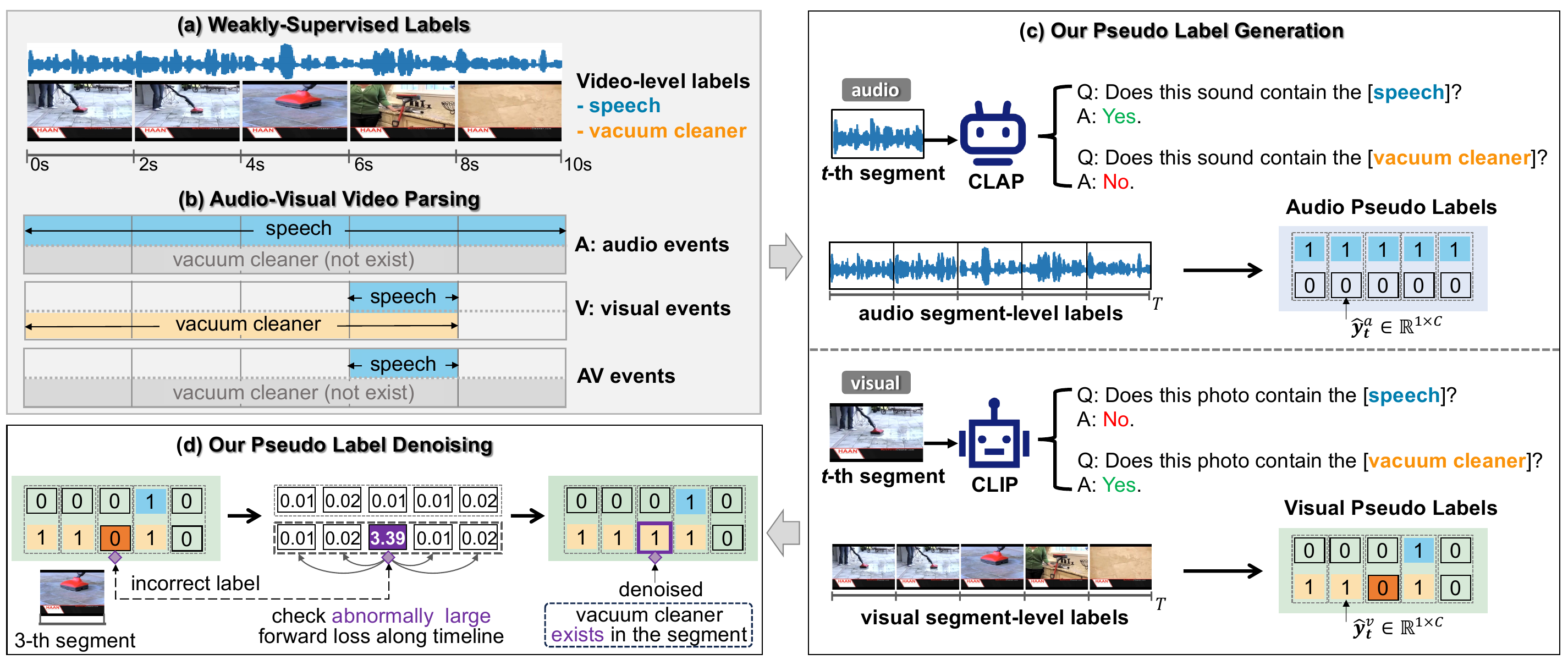}
  \caption{\textbf{An illustration of the weakly-supervised audio-visual video parsing (AVVP) task and our pseudo label exploration method.} 
  (a) Given a video and its event label (\textit{``speech"} and \textit{``vacuum cleaner"}),  (b) AVVP task needs to predict and localize the audio events, visual events, and audio-visual events.
  Note that \textit{"vacuum cleaner"} only exists in the visual track, while \textit{"speech"} exists in both audio and visual tracks, resulting in the audio-visual event \textit{"speech"}.
  (c) To ease this challenging weakly-supervised task, we aim to explicitly assign reliable segment-level audio and visual pseudo labels.
  In our pseudo label generation process, the pretrained CLAP and CLIP models are used to tell what events occur in each audio and visual segment, respectively.
  (d) We further propose a pseudo label denoising strategy to improve the obtained visual pseudo labels by examining those segments that have abnormally large forward loss values. In the example, visual event \textit{vacuum cleaner} at the third segment is assigned an incorrect pseudo label `0' and gets a large forward loss. Our pseudo-label denoising strategy further amends this, giving the accurate pseudo label `1'.  
  }
  \vspace{-2mm}
  \label{fig:task_illustration}
\end{figure*}

\section{Introduction}\label{sec:introduction}
Acoustic and visual signals flood our lives in abundance, and each signal may carry various events.
For example, we often see driving cars and pedestrians walking around on the street. Meanwhile, we can hear the beeping of the car horns and the sound of people talking.
Humans achieve such a comprehensive understanding of audio-visual events in large part thanks to the simultaneous use of their auditory and visual sensors.
To imitate this kind of intelligence for machines, many research works started from some fundamental tasks of single modality understanding, such as the audio classification~\citep{hershey2017vggish,kong2018audio,kumar2018knowledge,gong2021ast}, video classification~\citep{karpathy2014large,long2018attention,long2018multimodal,tran2019video}, and temporal 
action localization~\citep{zeng2019graph,chao2018rethinking,zhu2021enriching,gao2022fine}.
The audio classification task focuses on the recognition of the audio modality, while the video classification and temporal action localization tasks focus on the visual modality.
With the deepening of research, many works have further explored the multi-modal audio-visual perception~\citep{wei2022learning}, giving birth to tasks such as sound source localization~\citep{arandjelovic2017look,rouditchenko2019self,arandjelovic2018objects,senocak2018learning,hu2020discriminative,hu2019deep,qian2020multiple,zhao2018sound,afouras2020self,zhou2022avs,zhou2023avss,sun2023learning}, audio-visual event localization~\citep{tian2018audio,wu2019dual,xu2020MM,zhou2021psp,mahmud2022ave,rao2022dual,xia2022cross,wu2022span,zhou2022cpsp,wang2023context}, audio-visual video description~\citep{shen2023fine} and question answering~\citep{yun2021pano,li2022learning,yang2022avqa,song2022memorial,li2023object}.

Recently, Tian \etal~\citep{tian2020HAN} proposed a new multi-modal scene understanding task, namely Audio-Visual Video Parsing (AVVP).
Given an audible video, the AVVP task asks to identify what events occur in the audio and visual tracks and in which video segments these events occur.
Accordingly, the category and temporal boundary of each event are expected to be predicted for each modality.
Note that both the audio and visual tracks may contain multiple distinct events, and these events usually exist in different consecutive segments, it is labor-intensive to provide segment-level event labels for each modality with strong supervision.
The fact is that the AVVP is performed in a weakly-supervised setting where only the video label is provided during model training.
As the example shown in Fig.~\ref{fig:task_illustration} (a), we only know that this video contains the event set of \textit{speech} and \textit{vacuum cleaner}.
For each event, the model needs to judge whether it exists in the audio modality (audio event), visual modality (visual event), or both (audio-visual event), and locate the specific temporal segments, respectively.
Notably, as illustrated in Fig.~\ref{fig:task_illustration} (b), in the AVVP task, the audio-visual event is the intersection of the audio event and visual event, whereas the video label is the union of the audio event and visual event.

In this work, we emphasize there are two main challenges in the AVVP task.
 {\textbf{1) Cross-modal interference from the video label.}} 
As the example shown in Fig.~\ref{fig:task_illustration} (b), given the weakly-supervised video label, the audio and the visual track share the same supervision, \ie, \{\textit{speech, vacuum cleaner}\} together. However, the audio
and visual tracks contain distinct events. 
The \textit{vacuum cleaner} only exists in the visual modality.
Thus, during the model training process, the label \textit{vacuum cleaner} will interfere with the audio event parsing.
Similarly, the visual event parsing may also be interfered with the audio event label in other samples. 
\textbf{2) Temporal segment distinction.}
Assuming we successfully identify there is an event \textit{vacuum cleaner} in the visual modality,
it is still hard to distinguish which segments contain this event (segment level) under the weakly-supervised labels (video level).
These two challenges make the AVVP an intractable Multi-modal Multi-Instance Learning (MMIL) problem, namely distinguishing the events from both \textit{modality} and \textit{temporal} perspectives.

In the pioneer work~\citep{tian2020HAN}, a benchmark named Hybrid Attention Network (HAN) is proposed to encode the audio-visual features, which uses attentive pooling to aggregate the audio and visual features to predict events of the video.
The weak video label is used as the main supervision.
To address this task, they propose to obtain the pseudo labels for separate audio and visual modalities by processing the known video label with label smoothing~\citep{szegedy2016rethinking} technique.
Their experimental results indicate that generating pseudo labels for each modality brings significant benefits for supervising event parsing~\citep{tian2020HAN}. 
The subsequent studies diverge into two branches.
Most of them focus on \textit{designing effective networks} to implicitly aggregate the multi-modal features for prediction~\citep{mo2022MGN,pasi2022investigating,lamba2021cross,yu2022MMPyramid,lin2021CVCMS,jiang2022DHHN,gao2023collecting}, while using the video-level pseudo labels generated by HAN~\citep{tian2020HAN}.
In contrast, the other new works~\citep{wu2021MA,cheng2022JoMOLD} devote to \textit{generating better pseudo labels} for each modality based on the baseline backbone of HAN.
However, the generated pseudo label is denoised from the video label and limited to the video level which only indicates what events exist in each modality (modality perspective).
Therefore, it fails to address the second challenge because it remains difficult to distinguish which segments contain the event (temporal perspective).

To deal with the above-mentioned two challenges, our work starts with the intuition that can we explicitly generate pseudo labels for \textbf{each segment} of \textbf{each modality} to facilitate this MMIL task. 
This is inspired by two observations: 1) The AVVP models are expected to be well-guided with segment-level labels as such fine-grained labels can provide more explicit supervision information and directly fit the goal of the AVVP task (temporal perspective);
2) The audio and visual signals are processed with independent sensors for humans.
We can indeed annotate each modality, specifically for what we hear or see, by leveraging unimodal input (modality perspective).
To this end, we propose a \textbf{Visual-Audio Pseudo LAbel exploratioN (VAPLAN) method} that aims to generate high-quality segment-level pseudo labels for both visual modality and audio modality and further advances this weakly-supervised AVVP task.

To obtain the visual or audio pseudo labels, a natural idea is to borrow free knowledge from pretrained models for the image or audio classification. However, there is a category misalignment problem between the source and the target datasets using such a strategy. Take generating visual pseudo labels as an example, the models typically pretrained on the ImageNet~\citep{deng2009imagenet} would classify the instance in the AVVP task into \textit{predefined categories of the ImageNet}. However, the predicted category label may not exist in the target LLP dataset of the AVVP task, causing the category misalignment. 
Different from the traditional image classification models, vision-language pre-training~\citep{alayrac2022flamingo,jia2021scaling,radford2021CLIP} has attracted tremendous attention recently, which can flexibly classify images from an open-category vocabulary and show impressive zero-shot performance.
Among those works, Contrastive Language-Image Pretraining (CLIP)~\citep{radford2021CLIP} is a representative one.
Given an image, its potential category names are inserted into a predefined text prompt.
Then CLIP can score the categories according to the similarity between the encoded texts and the image features.
The category with a high similarity score is finally identified as the classification result.
Similar to the CLIP, in the audio community, the Contrastive Language-Audio Pretraining (CLAP)~\citep{wu2023clap} is trained on a large-scale corpus that incorporates the texts with the semantic-aligned audio.
With similar training and inference schemes, CLAP is able to perform audio classification in a zero-shot manner, and satisfactorily identify the category of a given audio from open-vocabulary too.

Inspired by such benefits of large-scale pretraining, we propose a \textbf{Pseudo Label Generation (PLG)} module that seeks guidance from the CLIP~\citep{radford2021CLIP} and CLAP~\citep{wu2023clap} to generate reliable segment-level visual and audio pseudo labels.
A simple illustration of PLG can be seen from Fig.~\ref{fig:task_illustration} (c).
Given all the potential event labels, CLIP/CLAP acting like an intelligent robot is asked to answer whether the event is contained in the given visual/audio segment.
In brief, the queried event categories with high cross-modal similarity scores that exceed the pre-set threshold $\tau_v$/$\tau_a$ are finally regarded as the visual/audio pseudo labels. 
This process can be applied to each video segment, so we can obtain segment-level pseudo labels.
We provide more implementation details in Sec.~\ref{sec:method_PLG}.
The generated pseudo labels are used to provide 
full supervision for each modality. 
Going a step further, we consider the generated pseudo labels may contain potential noise since the pseudo labels are non-manually annotated. Especially, some video instances can be challenging even for human annotators due to issues inherent in the collected videos, such as objects in the visual event being too small or obscured. 
As the example shown in Fig.~\ref{fig:task_illustration} (d), only part of the \textit{vacuum cleaner} is visible in the third segment.
PLG only uses the single frame to generate pseudo labels and fails to recognize the visual event \textit{vacuum cleaner} for this segment without contextual information, giving the incorrect pseudo label `0' for this category (denoted by brown box).
To alleviate such noise in pseudo labels generated by PLG, we further propose a \textbf{Pseudo Label Denoising (PLD)} strategy to re-examine the generated pseudo labels and amend the incorrect ones.
Samples with noisy labels are usually hard to learn and often get a large forward propagation loss~\citep{hu2021learning,kim2022large,huang2019o2u}.
In our work, the large loss comes from those data where the model is unable to give consistent predictions with the pseudo labels.
For the video example shown in Fig.~\ref{fig:task_illustration} (d), the third segment indeed suffers an abnormally large forward loss whereas the value is 3.39. 
Note that the values are almost zero for other segments in the same video which are assigned accurate labels.
This motivates us to perform a segment-wise denoising by checking the abnormally large forward loss along the timeline.
The segments with these controversial pseudo labels will be reassigned, providing a more accurate version.
More discussions and implementation details of PLD will be introduced in Sec.~\ref{sec:method_PLD}.

PLG and PLD enable the production of high-quality pseudo labels. 
Furthermore, we find that the obtained segment-level audio and visual pseudo labels contain rich information, indicating \textit{how many categories of events happen in each audio/visual segment} (category-richness) and \textit{how many audio/visual segments a certain category of the event exists in} (segment-richness).
Take the visual modality for example, as shown in Fig.~\ref{fig:task_illustration} (b), the video-level label indicates that there may be at most \textit{two} events in the visual track, \ie, the \textit{speech} and \textit{vacuum cleaner}.
In practice, only the fourth segment contains both \textit{two} events while the first segment contains only \textit{one} event, namely the {vacuum cleaner}.
Therefore, we can denote the visual category richness for the first and the fourth segments as 1/2 and 1, respectively.
Similarly, from the perspective of the event categories, the vacuum cleaner event appears in \textit{four} video segments of the entire video which totally contains \textit{five} segments, while the speech event only exists in \textit{one} (the fourth) segment.
Thus, we can denote the visual segment richness for events of \textit{vacuum cleaner} and \textit{speech} as 4/5 and 1/5, respectively.
Such information about category richness and segment richness can also be observed in the audio track. 
An AVVP model should be aware of the differences in category richness and segment richness to give correct predictions.
Based on this, we propose a \textbf{Pseudo Label Exploitation (PLE)} strategy that uses a novel \textit{Richness-aware Loss} to align the richness information contained in model predictions with that contained in pseudo labels.
Our experiments verify that the generated pseudo labels combined with the proposed richness-aware loss significantly boost the video parsing performance.

For the challenging audio-visual video parsing task, 
we conduct a comprehensive study on the exploration of the segment-wise audio and visual pseudo labels, including their generation, exploitation, and denoising.
Extensive experimental results demonstrate the effectiveness of our main designs.
Besides, our method can also be extended to the related weakly-supervised audio-visual event localization (AVEL)~\citep{tian2018audio,wu2019dual,zhou2021psp} task.
Overall, our contributions can be summarized as follows:
\begin{itemize}
    \item[$\bullet$] We introduce a new approach to explore the pseudo-label strategy for the AVVP task from a more fine-grained level, \ie, the segment level.
    \item[$\bullet$] Our proposed pseudo label generation and label denoising strategies successfully provide high-quality segment-wise audio and visual pseudo labels.
    \item[$\bullet$] We propose a new richness-aware loss function for superior model optimization, effectively exploiting the segment-richness and category-richness present in the pseudo labels.
    \item[$\bullet$] Our method achieves new state-of-the-art in all types of event parsing, including audio event, visual event, and audio-visual event parsing.
    \item[$\bullet$] The proposed core designs can be seamlessly integrated into existing frameworks for the AVVP task and AVEL task, leading to enhanced performances. 
\end{itemize}

\section{Related Work}\label{sec:related_work}

\textbf{Audio-Visual Video Parsing (AVVP).}
AVVP task needs to recognize what events happen in each modality and localize the corresponding video segments where the events exist.
Tian \etal~\citep{tian2020HAN} first propose this task and design a hybrid attention network to aggregate the intra-modal and inter-modal features.
Also, they use the label smoothing~\citep{szegedy2016rethinking} strategy to address the modality label bias from the single video-level label.
Some methods focus on network design.
Yu \etal ~\citep{yu2022MMPyramid} propose a multimodal pyramid attentional network that consists of multiple pyramid units to encode the temporal features. 
Jiang \etal ~\citep{jiang2022DHHN} 
use two extra independent visual and audio prediction networks to alleviate the label interference between audio and visual modalities.
Mo \etal ~\citep{mo2022MGN} use learnable class-aware tokens to group the semantics from separate audio and visual modalities. 
To overcome the label interference, Wu \etal ~\citep{wu2021MA} swap the audio and visual tracks of two event-independent videos to construct new data for model training.
The pseudo labels are generated according to the predictions of the reconstructed videos.
Cheng \etal ~\citep{cheng2022JoMOLD} first estimate the noise ratio of the video label and reverse a certain percentage of the label with large forward losses.
Although these methods bring considerable improvements, they can only generate the event label from the video level.
Unlikely, we aim to directly obtain high-quality pseudo labels for both audio and visual modalities from the segment level that further helps the video parsing system training.

\textbf{CLIP/CLAP Pre-Training.} 
Here, we discuss the pre-training technique and elaborate on 
why we choose the CLIP/CLAP as the base big model for generating pseudo labels in this work. CLIP~\citep{radford2021CLIP} is trained on a dataset with 400 million \textit{image-text} pairs using the contrastive learning technique.
This large-scale pretraining enables CLIP to learn efficient representations of the images and texts and demonstrates impressive performance on zero-shot image classification.
Its zero-shot transfer ability opens a new scheme to solve many tasks and spawns a large number of research works, such as image caption~\citep{Barraco_2022_CVPR}, video caption~\citep{tang2021clip4caption}, and semantic segmentation~\citep{ma2022open,ding2022decoupling,xu2021simple,zhou2022extract,rao2022denseclip}.
Most of the works choose to freeze or fine-tune the image and text encoders of CLIP to extract advanced features for downstream tasks~\citep{tang2021clip4caption,wang2022cris,Barraco_2022_CVPR,ma2022open,zhou2022zegclip}.
For the zero-shot semantic segmentation, some methods start to use the pretrained CLIP to generate pixel-level pseudo labels which are annotator-free and helpful~\citep{zhou2022extract,rao2022denseclip}.
Similarly to CLIP, CLAP~\citep{wu2023clap} is trained using a similar contrastive objective but with 630k \textit{audio-text} pairs and achieves state-of-the-art zero-shot audio classification performance.
Recently, some works have started to use CLAP to facilitate downstream tasks, such as audio source separation~\citep{liu2023separate}, text-to-audio generation~\citep{liu2023audioldm}, and speech emotion recognition~\citep{pan2023gemo}.
In this work, we make 
a new attempt to borrow the prior knowledge from CLIP/CLAP to ease the challenging weakly-supervised audio-visual video parsing task.

\textbf{Learning with Pseudo Labels}.
Deep neural networks achieve remarkable performance in various tasks, largely due to the large amount of labeled data available for training.
Recently, some researchers have attempted to generate massive pseudo labels for unlabeled data to further boost model performance. 
Most methods directly generate and use pseudo labels, which have been proven to be beneficial for various tasks, such as image classification~\citep{yalniz2019billion,xie2020self,pham2021meta,rizve2021defense,zoph2020rethinking,Hu_2021_CVPR}, speech recognition~\citep{kahn2020self,park2020improved}, and image-based text recognition~\citep{patel2023seq}. 
For the studied AVVP task, few works study the impact of pseudo labels and existing several methods focus on disentangling the event pseudo label for each modality from the known video label~\citep{tian2020HAN,wu2021MA,cheng2022JoMOLD}.
However, the obtained pseudo labels are confined to 
the \textit{video level}.
On the other hand, some new works in other fields notice the potential noise contained in the pseudo labels and propose effective methods to better learn with noisy pseudo labels~\citep{hu2021learning,kim2022large}.
Specifically, Hu \etal~\citep{hu2021learning} propose to optimize the network by giving much weight to the clean samples while less on the hard-to-learn samples. In the weakly-supervised multi-label classification problem, Kim \etal~\citep{kim2022large} propose to correct the false negative labels that are likely to have larger losses. 
However, these works focus on label refinement for image tasks. Refocusing on our video task, we conduct a comprehensive exploration of pseudo labels, encompassing both their generation and denoising.
Specifically, we propose to assign explicit pseudo labels for \emph{each segment of each modality}. We achieve this goal by flexibly sending all the possible event categories to reliable large-scale text-vision/audio models to pick the most likely event categories for each video segment. Furthermore, we propose a new pseudo-label denoising strategy, which performs \textit{segment-wise} denoising to provide pseudo labels with more accurate temporal boundaries within each video. We also provide more in-depth discussions on pseudo-label quality assessment and the denoising effects in different modalities as shown in Sec.~\ref{exp:parameter_study}.


\section{Preliminary}
\label{sec:pre}
In this section, we formulate the detail of the AVVP task and briefly introduce the baseline framework HAN~\citep{tian2020HAN}, which is used in both our approach and prior works employing video-level pseudo labels \citep{wu2021MA,cheng2022JoMOLD} 
in the AVVP task.

\textbf{Task Formulation.} 
Given a $T$-second video sequence $\{V_t, A_t\}_{t=1}^{T}$, $V_t$ and $A_t$ denote the visual and the audio components at the $t$-th video segment, respectively.
The event label of the video $ \bm{y}^{v \cup a}  \in \mathbb{R}^{1 \times C} = \{ y^{v \cup a}_c | y^{v \cup a}_c \in \{0, 1\}, c = 1, 2, ..., C\}$, 
where $C$ is the total number of event categories, the superscript `$v\cup a$' denotes the event label of the entire video is the union of the labels of audio and visual modalities, value 1 of ${y}^{v \cup a}_c$ represents an event with that $c$-th category happens in the video.
Note that $\bm{y}^{v \cup a}$ is a weakly-supervised label from the video level, \emph{the label of each individual modality for each video segment is unknown during training.
However, the audio events, visual events, and audio-visual events contained in each segment need to be predicted for evaluation.}
We denote the probabilities of the video-level visual and audio events as $\{\{\bm{p}^v; \bm{p}^a\} \in \mathbb{R}^{1 \times C} | {p}^v_c, {p}^a_c \in [0, 1] \}$,
$\bm{p}^{v \cap a}=\bm{p}^{v}*\bm{p}^{a}$ is used to represent the intersection of them.
Thus, the probability of the visual events, audio events, and audio-visual events of all video segments can be denoted as $\{\bm{P}^v; \bm{P}^a; \bm{P}^{v \cap a} \} \in \mathbb{R}^{T \times C} $, which need to be predicted.

\textbf{Baseline Framework.}
The baseline network HAN\\~\citep{tian2020HAN} uses the multi-head attention (\texttt{MHA}) mechanism in Transformer~\citep{vaswani2017attention} to encode intra-modal and cross-modal features for audio and visual modalities.
We denote the initial audio and visual features extracted by pretrained 
neural networks~\citep{hershey2017vggish,he2016resnet} as $ \bm{F}^a $, $\bm{F}^v  \in \mathbb{R}^{T \times d}$, where $d$ is the feature dimension.
The process of HAN can be summarized as,
\begin{equation}
\left\{
\begin{gathered}
\bm{\dot{F}}^a = \bm{F}^a + \texttt{MHA}(\bm{F}^a, \bm{F}^a) + \texttt{MHA}(\bm{F}^a, \bm{F}^v), \\
\bm{\dot{F}}^v = \bm{F}^v + \texttt{MHA}(\bm{F}^v, \bm{F}^v) + \texttt{MHA}(\bm{F}^v, \bm{F}^a),
\end{gathered}
\right.
\label{eq:HAN}
\end{equation}
where $\bm{\dot{F}^a}$, $\bm{\dot{F}^v} \in \mathbb{R}^{T \times d}$ are the updated audio and visual features.
The probabilities of segment-wise events for audio and visual modalities are predicted through a fully-connected (FC) layer and a sigmoid function, denoted as $\bm{P}^a \in \mathbb{R}^{T \times C}$ and $\bm{P}^v \in \mathbb{R}^{T \times C}$. 
An attentive pooling layer is further used to transform the segment-level predictions $\{ \bm{P}^a; \bm{P}^v\}$ to video-level predictions $\{ \bm{p}^a; \bm{p}^v\} \in \mathbb{R}^{1 \times C}$. 
By summarizing the audio and visual predictions, $\bm{p}^a$ and $\bm{p}^v$, we obtain the event prediction of the entire video $\bm{p}^{v \cup a} \in \mathbb{R}^{1 \times C}$. 
The basic video-level objective for model training is:
\begin{equation}\label{eq:HAN_loss}
\mathcal{L} = \mathcal{L}_{\text{bce}}(\bm{p}^{v \cup a}, \bm{y}^{v \cup a}) + \mathcal{L}_{\text{bce}}(\bm{p}^a, \overline{\bm{y}}^a) + \mathcal{L}_{\text{bce}}(\bm{p}^v, \overline{\bm{y}}^v),
\end{equation}
where $\mathcal{L}_{\text{bce}}$ is the binary cross-entropy loss, $\bm{y}^{v \cup a} \in \mathbb{R}^{1 \times C} $ is the video-level ground truth label and $\{\overline{\bm{y}}^v; \overline{\bm{y}}^a \} \in \mathbb{R}^{1 \times C} $ are the video-level visual and audio pseudo labels generated using label smoothing~\citep{szegedy2016rethinking} from $\bm{y}^{v \cup a}$.

\begin{figure*}[t]
\centering
\includegraphics[width=\textwidth]{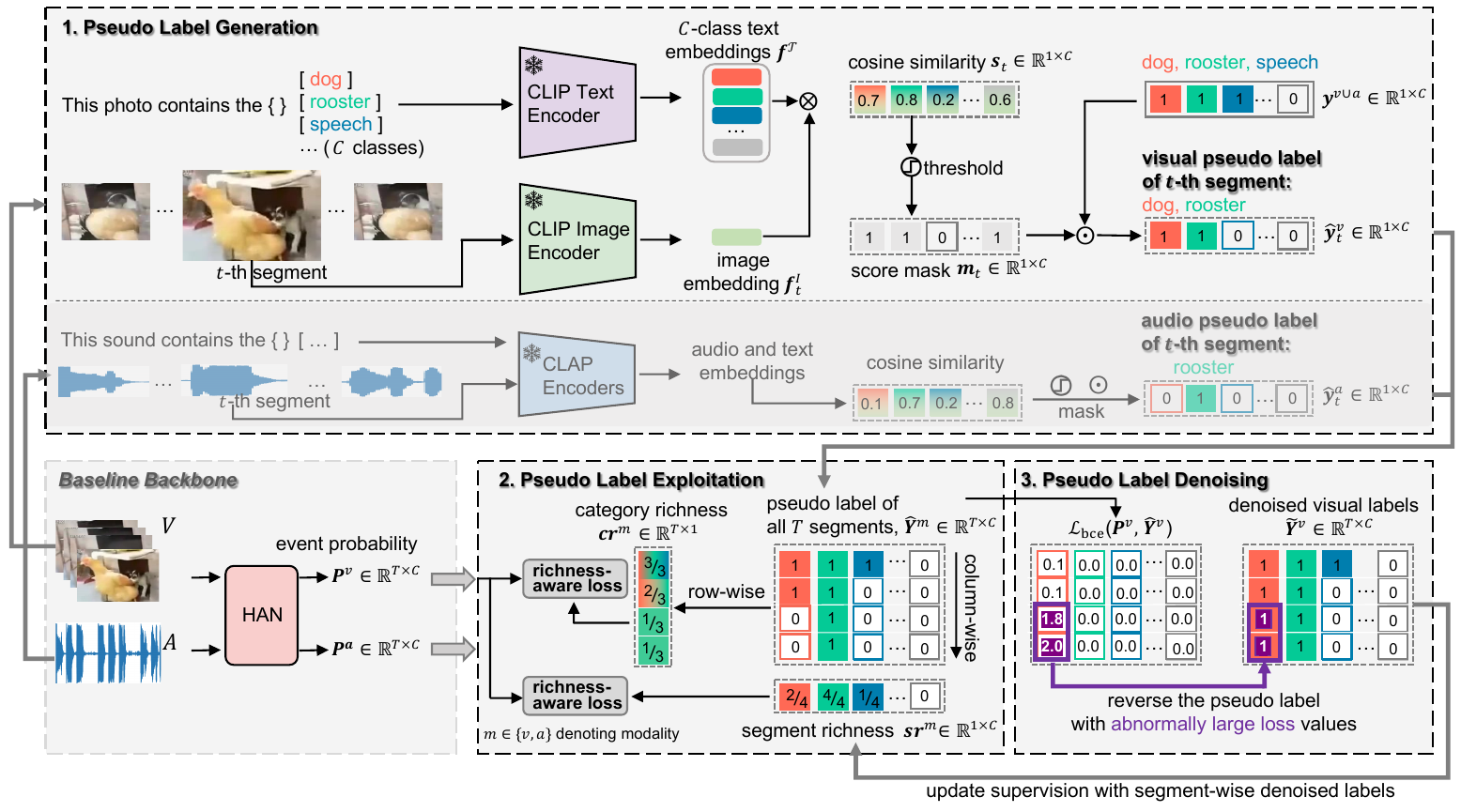}
\caption{\textbf{Overview of our method.} As a label refining method, we aim to produce high-quality and fine-grained segment-wise event labels. For the backbone, any existing network for the AVVP task can be used to generate event predictions. Here, we adopt the baseline HAN~\citep{tian2020HAN}.
In our solution, we design a \textit{pseudo label generation (PLG)} module, where the pretrained CLIP~\citep{radford2021CLIP} and CLAP~\citep{wu2023clap} are used to generate segment-level pseudo labels for the visual and the audio modality, respectively. 
Notably, the parameters of the CLIP and CLAP are frozen.
In the figure, we detail the visual pseudo label generation and simplify that for the audio modality since they share similar pipelines. 
In brief, the pseudo labels can be identified by thresholding the similarity of visual/audio--(event) text embeddings.
For the $t$-th segment, the video label \emph{`speech'} is filtered out for the visual modality and only 
\emph{`rooster'} is remained for the audio modality. 
After that, with the generated pseudo labels, we propose the \textit{pseudo label exploitation (PLE)} by designing a richness-aware loss as a new fully supervised objective to help the model align the category richness and segment richness in the prediction and pseudo label. 
Lastly, we design a \textit{pseudo label denoising (PLD)} strategy that further refines the pseudo labels by reversing the positions with anomalously large forward loss values. Specifically, we re-examine the pseudo labels along the timeline.
Pseudo labels of those segments with abnormal high binary cross-entropy forward loss will be refined (the motivation and implementation detail can be seen in Sec.~\ref{sec:method_PLD}).
The updated pseudo labels are further used as new supervision for model training. 
$\otimes$ denotes the matrix multiplication and $\odot$ is the element-wise multiplication.}
\label{fig:framework}
\end{figure*}

\section{Our Method}
An overview of our method is shown in Fig.~\ref{fig:framework}.
We focus on producing reliable segment-level audio and visual pseudo labels to better supervise the model for audio-visual video parsing.
For the backbone, we simply adopt the baseline HAN~\citep{tian2020HAN} to generate event predictions.
Our method provides the following new innovations. 
1) We propose a \textbf{pseudo label generation} module that uses the pretrained CLIP~\citep{radford2021CLIP} and CLAP~\citep{wu2023clap} models to respectively generate reliable visual and audio pseudo labels from the segment level.
2)  
We then propose a \textbf{pseudo label exploitation} strategy to utilize the obtained pseudo labels.
Specifically, we design a new \textit{richness-aware loss} to regularize the predictions to be aware of the category richness and segment richness contained in the pseudo labels.
This is helpful for model optimization.
3) We also propose a \textbf{pseudo label denoising} strategy that further improves the generated visual pseudo labels for those data with abnormally high forward loss values due to being assigned incorrect pseudo labels.
Next, we elaborate on these proposed strategies.

\subsection{Pseudo Label Generation (PLG)}\label{sec:method_PLG}
PLG aims to generate high-quality visual and audio pseudo labels from the segment level that are expected to alleviate the video-level label interference 
for single modality and better supervise the model to distinguish video segments. 
As discussed in Sec.~\ref{sec:introduction}, we select the pretrained CLIP~\citep{radford2021CLIP} and CLAP~\citep{wu2023clap} to achieve this goal due to their flexible open-vocabulary classification capabilities. 

Taking visual modality as an example, we detail the pseudo label generation process.
Specifically, each video instance is evenly split into several segments and we sample the middle frame to represent each segment.
As shown in Fig.~\ref{fig:framework}-1, for the sampled frame $\mathcal{I}_t$ at the $t$-th segment, we input it into CLIP image encoder and obtain 
the visual feature, denoted as $\bm{f}^{\mathcal{I}}_t \in \mathbb{R}^{1 \times d}$. As for the event category encoding, the default text input of the CLIP text encoder follows the prompt ``\texttt{A photo of a [CLS]}'' where the \texttt{[CLS]} can be replaced by the potential category names. For the AVVP task, we empirically change the prompt to a more appropriate one, ``\texttt{This photo contains the [CLS]}'' (An ablation study of prompt in CLIP text encoder will be shown in Sec.~\ref{exp:parameter_study}). 
By replacing the \texttt{[CLS]} in this prompt with each event category and sending the generated texts 
to the CLIP text encoder, we can obtain the text (with event category) features of all $C$-class $\bm{f}^{\mathcal{T}} \in \mathbb{R}^{C \times d}$. 
Then the normalized cosine similarity $\bm{s}_t \in \mathbb{R}^{1 \times C}$ between the image and event categories can be computed by,
\begin{equation}\label{eq:cosine_simmlarity}
\bm{s}_t = softmax(\frac{ \bm{f}^{\mathcal{I}}_t }{\| \bm{f}^{\mathcal{I}}_t \|_2 }  \otimes (\frac{ \bm{f}^{\mathcal{T}}}{ \| \bm{f}^{\mathcal{T}} \|_2})^{\top}),
\end{equation}
where $\otimes$ denotes the matrix multiplication, and $\top$ is the matrix transposition. A high similarity score in $\bm{s}_t$ indicates that the event category is more likely to appear in the image.

We use a threshold $\tau_v$ to select the categories with higher confidence scores in $\bm{s}_t$ and obtain the score mask $\bm{m}_t$. After that, we impose the score mask $\bm{m}_t$ on the known video-level label $\bm{y}^{v \cup a}$ with element-wise multiplication $\odot$ to filter out the visual events occurring at $t$-th segment $\bm{\hat{y}}^v_t \in \mathbb{R}^{1 \times C}$. This process can be formulated as,
\begin{equation}\label{eq:PLG_tau}
\left\{
\begin{gathered}
\begin{split}
    & \bm{m}_t = \mathbbm{1}({\bm{s}_t - \tau_v}), \\
    & \bm{\hat{y}}^v_t = \bm{m}_t \odot \bm{y}^{v \cup a},
\end{split}
\end{gathered}
\right.
\end{equation}
where $\mathbbm{1}(x_i)$ outputs `1' when the input $x_i \geq 0$ else outputs `0', $ i=1, 2, ..., C$, and $\bm{m}_t \in \mathbb{R}^{1 \times C}$. 

This pseudo label generation process can be applied to all the segments. Therefore, we can obtain the segment-level visual pseudo label for each video, denoted as $\bm{\hat{Y}}^v =\{ \bm{\hat{y}}^v_t \}\in \mathbb{R}^{T \times C}$.  
Note that the video-level visual pseudo label $\hat{\bm{y}}^v \in \mathbb{R}^{1 \times C}$ can be easily obtained from $\bm{\hat{Y}}^v$, where $ \hat{{y}}^v_c =  \mathbbm{1}(\sum_{t=1}^{T} \bm{\hat{Y}}^v_{t,c})$ that means if a category of the event exists in at least one video segment, it is contained in the video-level label.

As for the audio pseudo labels, they can be generated in a similar way 
but with several adjustments.
For brevity, we introduce the main steps here. 1) We use the CLAP model instead of the CLIP for audio pseudo label generation. 2) The audio waveform of the entire video is split into $T$ equal-length segments and each segment is sent to the CLAP audio encoder.
3) We use the prompt ``\texttt{This sound contains the [CLS]}'' with the event categories as the input of CLAP text encoder.
4) We compute the similarity score of the text and audio features extracted by CLAP (just like Eq.~\ref{eq:cosine_simmlarity}) and use an independent threshold $\tau_a$ (replace $\tau_v$ in Eq.~\ref{eq:PLG_tau}) to select high similarity values.
In this way, 
we obtain the segment-level audio pseudo label $\bm{\hat{Y}}^a \in \mathbb{R}^{T \times C}$ and the video-level audio pseudo label $\hat{\bm{y}}^a \in \mathbb{R}^{1 \times C}$ for each video sample.

\subsection{Pseudo Label Exploitation (PLE)} 

The weakly-supervised AVVP task requires predicting for each segment, but only the video-level label is provided. 
This task would be greatly advanced if segment-level supervision is additionally provided.
In this part, we try to exploit the pseudo labels 
from both the video-level and segment-level since we have obtained pseudo labels of these two levels, namely $\hat{\bm{y}}^m \in \mathbb{R}^{1 \times C}$ and $\hat{\bm{Y}}^m \in \mathbb{R}^{T \times C}$, where $m \in \{v, a\}$ denotes the modality type. 
In particular, for the segment-level supervision, we propose a new richness-aware optimization objective to help the model align the predictions and pseudo labels.
We introduce our pseudo label exploitation strategy in the two aspects below.

\textbf{Basic video-level loss.} Existing methods usually adopt the objective function formulated in Eq.~\ref{eq:HAN} for model training~\citep{wu2021MA,yu2022MMPyramid,cheng2022JoMOLD,mo2022MGN}, where $\overline{\bm{y}}^m \in \mathbb{R}^{1 \times C}$ is the video-level label obtained by label smoothing.
Instead, we use the video-level pseudo label $\hat{\bm{y}}^m \in \mathbb{R}^{1 \times C}$ generated by our PLG module as new supervision. The objective is then updated to,
\begin{equation}\label{eq:RL-video_level}
\mathcal{L_V} = \mathcal{L}_{\text{bce}}(\bm{p}^{v \cup a}, \bm{y}^{v \cup a}) + \mathcal{L}_{\text{bce}}(\bm{p}^a, \hat{\bm{y}}^a) + \mathcal{L}_{\text{bce}}(\bm{p}^v, \hat{\bm{y}}^v).
\end{equation}

\textbf{New segment-level loss.} With the segment-wise pseudo label $\hat{\bm{Y}}^m$, we propose a new richness-aware loss that is inspired by the following observations. \textbf{1)}
Each row of the segment-wise pseudo labels, \eg, $\hat{\bm{Y}}^m_{t\cdot} \in \mathbb{R}^{1 \times C}$, the $t$-th row of the pseudo label,
indicates whether all the events appear in the $t$-th segment.
For example, we show the visual pseudo label in Fig.~\ref{fig:framework}-2, \ie, $\hat{\bm{Y}}^m$ where $m=v$.
There are three visual events in the first segment, \ie, the \textit{dog}, \textit{rooster}, and \textit{speech}, $\hat{\bm{Y}}^v_{1\cdot} = [1,1,1]$, while 
the last segment only contains one \textit{rooster} event, \ie, $\hat{\bm{Y}}^v_{T\cdot} = [0,1,0]$. This reflects the richness of the event category in different segments that indicates \textit{how many event categories exist in each segment}.
Similarly, the audio pseudo label $\hat{\bm{Y}}^a$ tells the category richness of audio events.
We define the \textbf{category richness} of $t$-th segment ${cr}_t^m$ as the ratio of the category number of $t$-th segment to the total category number of the video, written as, 
\begin{equation}
   {cr}_t^m = \frac{ \sum_{c=1}^C \hat{\bm{Y}}^m_{t,c} }{ \sum_{c=1}^C \bm{y}^{v \cup a}_{c}},
   \label{eq:category_richness}
\end{equation}
where $m \in \{v, a\}$ denotes the visual or audio modality.
Therefore, we can obtain the category richness vector of all segments $\bm{cr}^m \in \mathbb{R}^{T \times 1}$ for each modality. In the example shown in Fig.~\ref{fig:framework}-2, the visual category richness for the first and last segments, \ie, $cr_1^v$ and  $cr_T^v$, is equal to 1 and 1/3, respectively.

\textbf{2)} On the other hand, each column of the pseudo labels, \eg, $\hat{\bm{Y}}^m_{\cdot c}\in \mathbb{R}^{T \times 1}$, $m \in \{v, a\}$, indicates \textit{how many visual/audio segments contain the event of $c$-th category.}
We denote the \textbf{segment richness} of $c$-th category ${sr}_c^m$ as the ratio of the number of segments containing that category $c$ to the total segment number of the video, 
written as below, 
\begin{equation}
   {sr}_c^m = \frac{1}{T} \sum_{t=1}^T \hat{\bm{Y}}^m_{t,c}.
   \label{eq:segment_richness}
\end{equation}
In the example shown in Fig.~\ref{fig:framework}-2, the visual segment richness for the event categories \textit{dog} and \textit{speech}, \ie, $sr_1^v$ and $sr_3^v$ is equal to 1/2 and 1/4, respectively. 
Extending to all $C$ event categories, we can obtain the segment richness vector of all the categories $\bm{sr}^m \in \mathbb{R}^{1 \times C}$, where $m \in \{v, a\}$ denotes the visual and audio modalities.

So far, regardless of modality $m \in \{v, a\}$, 
we can obtain the category richness $\bm{cr}^m$ and segment richness $\bm{sr}^m$ of the pseudo label. 
With the prediction $\bm{P}^m \in \mathbb{R}^{T \times C}$ from the baseline network, we can compute its category richness and segment richness in the same way, 
denoted as $\bm{pcr}^m \in \mathbb{R}^{T \times 1}$ and $\bm{psr}^m \in \mathbb{R}^{1 \times C}$. 
Then, we design the segment-level richness-aware loss $\mathcal{L_S}$ 
to align the richness of the predictions and the pseudo labels, calculated by,
\begin{equation}\label{eq:RL-segment_level}
  \mathcal{L_S} = \sum_{m\in\{v, a\}} {\mathcal{L}_{\text{bce}}(\bm{pcr}^m, \bm{cr}^m) + \mathcal{L}_{\text{bce}}(\bm{psr}^m, \bm{sr}^m)}.
\end{equation}

The total objective function $\mathcal{L}_{\text{total}}$ for AVVP in this work is the combination of the 
basic loss $\mathcal{L_V}$ and the 
richness-aware loss $\mathcal{L_S}$, \ie,
\begin{equation}\label{eq:RL}
    \mathcal{L}_{\text{total}} = \mathcal{L_V} + \lambda \mathcal{L_S},
\end{equation}
where $\lambda$ is a weight parameter.

\subsection{Pseudo Label Denoising (PLD)}
\label{sec:method_PLD}
In general, PLG can produce trustworthy segment-level pseudo labels, especially when combined with the proposed richness-aware loss, which significantly improves the audio-visual video parsing performance.
This can be verified by our experiments shown in Sec.~\ref{exp:ablations}. 
Going a step further, we posit that the generated pseudo labels may still encompass some noise. 
By our observation, the video-level event category pseudo-annotation can be satisfactorily tackled, but the misclassification of specific segments exists along the timeline within each video, particularly when dealing with hard video instances that are difficult to annotate from the segment level.
We specifically trace such challenges in the visual modality
and observe that without contextual information, separate frames sent to the CLIP may be incorrectly classified, especially in the instances where the visual objects in the images are too diminutive, the images are afflicted by blurriness or inadequate lighting, when portions of the objects are obscured, rendering them arduous to discern, \etc.
As shown in Fig.~\ref{fig:framework}, the \textit{dog} at the last two segments is mostly obscured by the \textit{rooster}, and CLIP fails to recognize the visual event \textit{dog} without contextual information.
In this case, the generated pseudo labels do not accurately capture the temporal boundary of the event and would be detrimental to model training. We believe that the segment-level visual pseudo labels can be further refined. 
As for the audio modality, the audio signal is represented through waveform and it keeps good continuity even if it is split into multiple segments for pseudo-labeling.
This characteristic may help to resist disturbances along the timeline when generating segment-level audio pseudo labels with CLAP. 
In fact, the quality of audio pseudo labels is indeed better than that of visual pseudo labels.
For example, the segment-level F-score metric for audio pseudo labels is $\sim$10 points higher than that of visual pseudo labels, as demonstrated in Tables~\ref{tab:parameter_study_PLG},~\ref{tab:parameter_study_APLG}. This implies the high quality of audio pseudo labels produced by PLG and highlights the greater difficulty in enhancing the accuracy of visual pseudo labels.
We present further discussions with more experimental results in Sec.~\ref{exp:parameter_study}.

In this section, we propose a pseudo label denoising (PLD) strategy that aims to recheck the pseudo labels generated by PLG and further refine the inaccurate ones (noisy pseudo labels).
Our PLD is inspired by the works that conduct label denoising with the help of the forward propagation loss 
for image tasks~\citep{kim2022large,hu2021learning}. 
In general, a large forward loss means that the trained model does not give the same prediction as the labels for a sample.
There are two main reasons for this: 1) the provided label is correct but the video data is hard to learn and the model does not learn an effective representation for it; 2) the label itself is incorrect.
In this work, our PLD aims to leverage the forward loss to check the temporal continuity of segment-level pseudo labels in each video and amend the abnormal segments when they belong to the second case.

Specifically, we first use the objective function shown in Eq.~\ref{eq:RL} to train a baseline model.
Then, we compute the element-wise forward loss matrix 
by measuring the binary cross entropy between the prediction $\bm{P}^m$ and the pseudo label $\bm{\hat{Y}}^m$, denoted as $ \bm{\mathcal{M}}^m = \mathcal{L}_{\text{bce}}(\bm{P}^m, \bm{\hat{Y}}^m) \in \mathbb{R}^{T \times C}$, where $m\in\{v,a\}$ denotes the visual and the audio modality. 
Denote the $j$-th column of $\bm{\mathcal{M}}^m$ as $\bm{\mathcal{M}}_{\cdot j}^m \in \mathbb{R}^{T \times 1}$, it indicates the loss value of all segments for the specific $j$-th event category.
In the example shown in Fig.~\ref{fig:framework}-3, we display the forward loss matrix for the visual modality and find that the last two video segments have much larger forward losses than other segments for the \textit{dog} category; they actually contain this event like other segments.
The abnormally large loss value is caused by the fact that the last two segments are assigned incorrect visual pseudo labels.
Therefore, the 
matrix $\bm{\mathcal{M}}^m$ can reflect those segments whose pseudo labels contain potential noise and require refinement.

Note that the pseudo label $\bm{\hat{y}}^m \in \mathbb{R}^{1 \times C}$ indicates the predicted event categories that appear in each modality.
We trust the event category $\bm{\hat{y}}^m$ and use it to mask the matrix $\bm{\mathcal{M}}^m$. 
There are two steps for the matrix $\bm{\mathcal{M}}^m$ masking. \textbf{Step I:} For other event categories that do not occur in the video sample, their pseudo labels will be eased by setting zeros in $\bm{\mathcal{M}}^m$. 
For the example shown in Fig.~\ref{fig:framework}-2, we only need to denoise the pseudo labels for the three columns of $\bm{\hat{y}}^m$ that corresponds to the predicted event categories of \textit{dog}, \textit{rooster} and \textit{speech}. 
The calculation of the masked 
matrix $\bm{\mathcal{M}'}^m$ can be computed by, 
\begin{equation}\label{eq:fw_loss}
\begin{gathered}
\begin{split}
    & \bm{\mathcal{M}'}^m = f_{\text{rpt-T}}(\bm{\hat{y}}^m) \odot \bm{\mathcal{M}}^m,
\end{split}
\end{gathered}
\end{equation}
where $\bm{\mathcal{M}'}^m \in \mathbb{R}^{T \times C}$, and $f_{\text{rpt-T}}(\bm{\hat{y}}^m)$ denotes the operation of repeating $\bm{\hat{y}}^m$ along the temporal dimension for $T$ times, and $f_{\text{rpt-T}}(\bm{\hat{y}}^m) \in \mathbb{R}^{T \times C}$.

\textbf{Step II:} Returning to the masked forward loss of all video segments of the $j$-th category $\bm{\mathcal{M}'}^m_{\cdot j} \in \mathbb{R}^{T \times 1}$,  we treat the average of the top-$K$ smallest loss values of $\bm{\mathcal{M}'}^{m}_{\cdot j}$ as the threshold $\mu_j^m$. 
$\mu_j^m$ is the tolerable forward loss within a video sample.
If the loss of some segments is abnormally larger than $\mu_j^m$, they may have incorrect pseudo labels.
Comparing the forward loss of each segment with $\mu_j^m$, we can obtain a binary mask vector $\bm{\varphi}_j^m \in \mathbb{R}^{T \times 1}$, where `1' reflects that the segment has a larger loss than $\mu_j^m$.
This process can be written as,
\begin{equation}
\left\{
\begin{gathered}
\begin{split}
    & \mu_j^m = f_{\text{avg}}(f_{\Bbbk}(\bm{\mathcal{M}'}^m_{\cdot j})), \\
     & \bm{\varphi}_j^m = \mathbbm{1}(\bm{\mathcal{M}'}^m_{\cdot j} - \alpha \cdot\mu_j^m),
\end{split}
\end{gathered}
\right.
\label{eq:PLD}
\end{equation}
where $f_{\Bbbk}$ and $f_{\text{avg}}$ denotes the top-$K$ minimum loss selection and the average operation, respectively. Note that we set a scaling factor $\alpha$ to magnify the averaged loss.
It is used to better ensure that anomalous loss is caused by incorrect pseudo labels rather than the data not being well learned.

Extending Eq.~\ref{eq:PLD} to all the event categories, we obtain the binary mask matrix of the video $\bm{\Phi}^m = \{ \bm{\varphi}_j^m \} \in \mathbb{R}^{T \times C}$. 
Afterwards, the segment-level pseudo label $\bm{\hat{Y}}^m$ produced by PLG can be refined by reversing the positions that have unusually large loss values reflected by $\bm{\Phi}^m$, denoted as 
$ \bm{\widetilde{Y}}^m = f_{\sim}(\bm{\hat{Y}}^m, \bm{\Phi}^m)$. 
As shown in Fig.~\ref{fig:framework}-2, for the event \textit{dog} again, the visual pseudo labels generated by PLG are `0' for the last two segments (indicating that there is no \textit{dog}) 
and get a large forward loss (marked by the purple box in Fig.~\ref{fig:framework}-3). 
This indicates that the visual pseudo labels of these two segments are incorrect (actually containing \textit{dog}) and are thus reversed during the denoising process.
We display more examples in Fig.~\ref{fig:vis_pseudo_labels} to illustrate the pseudo label denoising process.  
Finally, the pseudo labels refined by PLD can be taken as new supervision for the model training.

\section{Experiments}
\subsection{Experimental Setup}\label{exp:setup}
\textbf{Dataset.}
Experiments for the AVVP task are conducted on the publicly available \textit{Look, Listen, and Parse (LLP)}~\citep{tian2020HAN} dataset.
It contains 11,849 videos spanning over 25 common audio-visual categories, involving scenes such as humans, animals, vehicles, musical instruments, \etc.
Each video is 10 seconds long and around 61\% of the videos contain more than one event category.
Videos of the LLP dataset are split into 10,000 for training, 649 for validation, and 1,200 for testing.
The training set is provided with only the video-level labels, \ie, the label union of the audio events and visual events.
For validation and test sets, the segment-wise event labels for each audio and visual modality are additionally provided.

\textbf{Evaluation metrics.}
Following existing works~\citep{tian2020HAN,cheng2022JoMOLD,wu2021MA,yu2022MMPyramid}, we evaluate our method by measuring the parsing results of all the types of events, namely audio events (\textbf{A}), visual events (\textbf{V}), and audio-visual events (\textbf{AV}, both audible and visible). The average parsing result of the three types 
is denoted as the ``\textbf{Type@AV}'' metric. 
Different from \textbf{Type@AV} metric, ``\textbf{Event@AV}'' metric calculates the F-score considering the predictions of the audio and the visual events together.
For the above event types, both the segment-level and event-level F-scores are used as evaluation metrics.
The segment-level metric measures the quality of the predicted events by comparing them with the ground truth for each video segment.
And the event-level metric treats consecutive segments containing the same event category as a whole event, and computes the F-score based on mIoU = 0.5 as the threshold.
Therefore, the event-level F-score metric is more difficult because it requires the model to predict a satisfactory temporal boundary of the event.

\textbf{Implementation details.} \textit{1) Feature extraction.} For the LLP dataset,  each video is divided into 10 consecutive 1-second segments.
For a fair comparison, we adopt the same feature extractors to extract the audio and visual features.
Specifically, the VGGish~\citep{hershey2017vggish} network pretrained on AudioSet \citep{gemmeke2017audioset} dataset is used to extract the 128-dim audio features.
The pretrained ResNet152~\citep{he2016resnet} and R(2+1)D~\citep{tran2018closer} are used to extract the 2D and 3D visual features, respectively.
The low-level visual feature is the concatenation of 2D and 3D visual features. 
\textit{2) Pseudo label preparation}.
For each video in the training set of the LLP dataset, we first offline generate the segment-wise visual and audio pseudo labels using our PLG module. We use the ViT-B/32-based CLIP~\citep{vaswani2017attention} and HTSAT-RoBERTa-based CLAP~\citep{wu2023clap} to conduct the pseudo label generation, and their parameters are frozen. 
\textit{3) Training procedure}. 
The objective function $\mathcal{L}_{\text{total}}$ shown in Eq.~\ref{eq:RL} is used to train the baseline model HAN~\citep{tian2020HAN}.
The hyperparameter $\lambda$ in Eq.~\ref{eq:RL} for balancing the video-level and the segment-level losses is empirically set to 0.5. 
This pretrained model is then used in our PLD to further refine the pseudo labels.
The refined pseudo labels are used to supervise the baseline model training again.
For all the training processes, we adopt the Adam optimizer to train the model with a mini-batch size of 32 and the learning rate of $3 \times 10^{-4}$. The total training epoch is set to 30.
All experiments are conducted with PyTorch~\citep{paszke2019pytorch} on a single NVIDIA GeForce-RTX-2080-Ti GPU.
The codes, pseudo labels, and pretrained models will be released. 

\subsection{Parameter Studies}\label{exp:parameter_study}
We perform parameter studies of essential parameters used in our method, namely the score threshold $\tau_v$/$\tau_a$ and the text prompt for CLIP/CLAP used in the PLG module, and the top-$K$ and scaling factor $\alpha$ used in the PLD strategy.
Experiments in this section are conducted on the validation set of the LLP dataset of which the segment-level event labels are accessible. 
Thus, we also verify the quality of pseudo labels through \textit{correctness measurements} in this part.

\begin{table}[t]\small
\caption{\textbf{Parameter study of the threshold $\tau_v$ and prompt used in the {VISUAL} pseudo label generation.}
Different setups are used to generate segment-level pseudo labels; consequently, we can obtain the corresponding video-level pseudo labels. Here, we report the precision between the visual pseudo label and the ground truth from the video level.
Also, we report the segment-level and event-level F-scores.
\textit{`-' denotes the result of directly assigning video labels as the visual event labels and each event happens at all the visual segments.} 
The specific expressions of the prompts are introduced in our main text.
This experiment is conducted on the validation set of the LLP dataset.} 
\centering
\label{tab:parameter_study_PLG}
\small
\resizebox{\linewidth}{!}{
 \begin{tabular}{p{.6cm}<{\centering}p{1cm}<{\centering}p{.8cm}<{\centering}p{1.cm}<{\centering}p{1.cm}<{\centering}}
\toprule[0.8pt]
\multicolumn{2}{c}{Parameter setup} & \multicolumn{1}{c}{\multirow{2}{*}{Precision}}  & \multicolumn{1}{c}{\multirow{2}{*}{Segment. (V)}}  & \multicolumn{1}{c}{\multirow{2}{*}{Event. (V)}}  \\ \cmidrule{1-2}
\multicolumn{1}{c}{$\tau_v$} & \multicolumn{1}{c}{prompt} & & & \\ \midrule
- & - & 66.96 & 58.65 & 53.48 \\
\midrule
{0.040} & \multirow{3}{*}{VP1} & 85.31 & 70.29 & 64.68 \\
\textbf{0.041} & & \textbf{86.88} & \textbf{71.08} & \textbf{64.82} \\
0.042 &  & 72.19 & {51.51} & 43.13 \\ \midrule
\multirow{4}{*}{0.041} & \textbf{VP1} & \textbf{86.88} & \textbf{71.08} & \textbf{64.82}  \\
 & VP2 & 85.64 & 68.96 & 61.83 \\
 & VP3 & 84.69 & 67.60 & 60.98 \\
 & VP4 & {86.75} & 70.29 & 63.78 \\
\toprule[0.8pt]
\end{tabular}
}
\end{table}

\textbf{Study of the thresholds and prompts in PLG.}
$\tau_v$/$\tau_a$ is the threshold to select high scores of the cosine similarity between the event category and the visual/audio segment in the mask calculation (Eq.~\ref{eq:PLG_tau}). 
We first explore the impact of $\tau_v$ on the \textbf{visual pseudo label generation}. 
As shown in the upper part of Table~\ref{tab:parameter_study_PLG}, we used the default prompt \textbf{VP1} -- ``\texttt{This photo contains the [CLS]}'' and test several values of $\tau_v$ to generate visual pseudo labels.
\textit{Then, we report the category precision between the pseudo labels and the ground truth at the video level, and the segment-level and event-level F-scores to measure the quality of the generated pseudo labels.}
As shown in the Table, the pseudo label with the best quality is obtained when $\tau_v = $ 0.041.
And all the evaluation metrics drop significantly when $\tau_v$ changes from 0.041 to 0.042. We argue such sensitivity is related to the \emph{softmax} operation in Eq.~\ref{eq:cosine_simmlarity} that squeezes the similarity score into small logits. The metrics for visual modality are acceptable up to the threshold of $\tau_v$ =0.041.
Using the same experimental strategy, we explore the impacts of threshold $\tau_a$ in \textbf{audio pseudo label generation}.
The experimental results are shown in Table~\ref{tab:parameter_study_APLG} and we find that the optimal audio pseudo labels are obtained when $\tau_a$ is equal to 0.038.

\begin{table}[t]\small
\caption{\textbf{Parameter study of the threshold $\tau_a$ and prompt used in the AUDIO pseudo label generation.}
Different setups are used to generate segment-level audio pseudo labels.
Here, we report the segment-level and event-level F-scores between the audio pseudo label and the ground truth.
The last column shows the average value of these two evaluation metrics, which is used to select the best setup.
\textit{`-' denotes the result of directly treating the video labels as the audio event labels and each event happens at all the audio segments.}
The specific expressions of the prompts are introduced in our main text.
This experiment is conducted on the validation set of the LLP dataset.} 
\centering
\label{tab:parameter_study_APLG}
\small
\resizebox{\linewidth}{!}{
 \begin{tabular}{p{.6cm}<{\centering}p{1cm}<{\centering}p{1.cm}<{\centering}p{1.cm}<{\centering}p{1.cm}<{\centering}}
\toprule[0.8pt]
\multicolumn{2}{c}{Parameter setup} & \multicolumn{1}{c}{\multirow{2}{*}{Segment. (A)}}  & \multicolumn{1}{c}{\multirow{2}{*}{Event. (A)}}  & \multicolumn{1}{c}{\multirow{2}{*}{Average}}  \\ \cmidrule{1-2}
\multicolumn{1}{c}{$\tau_a$} & \multicolumn{1}{c}{prompt} & & & \\ \midrule
- & - & 77.07 & 63.84 & 70.45 \\
\midrule
0.037 & \multirow{4}{*}{AP1} & 79.79 & 70.77 & 75.28 \\
{0.038} & & 80.01 & 70.87 & {75.28} \\
{0.039} & & 80.23 & {71.27} & {75.75} \\
0.040 &  & 80.18 & {71.70} & 75.44 \\ \midrule
0.037 & \multirow{4}{*}{\textbf{AP2}} & 80.06 & 70.74 & 75.40 \\
\textbf{0.038} & & \textbf{80.32} & \textbf{71.54} & \textbf{75.93} \\
{0.039} & & 80.20 & {71.00} & {75.60} \\
0.040 &  & 80.03 & {69.91} & 74.97 \\ 
\toprule[0.8pt]
\end{tabular}}
\end{table}

Furthermore, we explore the impact of prompts used in the PLG. The prompts are combined with the event categories and sent as text inputs to the CLIP or CLAP text encoder. \textbf{ For the visual pseudo label generation}, specifically, we test four types of prompts, \ie, our default \textbf{VP1} -- ``\texttt{This photo contains the [CLS]}'', \textbf{VP2} -- ``\texttt{This photo contains the scene of [CLS]}'', \textbf{VP3} -- ``\texttt{This photo contains the visual scene of [CLS]}'' and \textbf{VP4} -- ``\texttt{This is a photo of the [CLS]}''.
We use these different prompts to generate pseudo labels and compare them with the ground truth. 
As shown in the lower part of Table~\ref{tab:parameter_study_PLG}, visual pseudo labels generated using these different prompts remain relatively consistent.
The pseudo label has the highest F-score using the \textbf{VP1} prompt.
Therefore, we 
use the prompt \textbf{VP1} as the default setup for visual pseudo label generation in our following experiments.
Notably, the precision of the video-level visual pseudo label reaches about 87\% under the optimal setup, whereas the precision of directly assigning video labels as the visual event labels (\ie, without prompt) is only $\sim$67\%. 
This reveals that PLG can satisfactorily disentangle visual events from weak video labels. 
\textbf{For the audio pseudo label generation}, we test 
two prompts, \ie, the \textbf{AP1} -- ``\texttt{This is a sound of [CLS]}'' and \textbf{AP2} -- ``\texttt{This sound contains the [CLS]}'', to generate segment level audio pseudo labels.
Then, \textit{we report the segment-level and event-level F-scores of the audio events} under different setups and use their average value to select the best one.
As shown in the Table~\ref{tab:parameter_study_APLG}, performances moderately change under different setups, and the best performance is obtained when using the \textbf{AP2} prompt and $\tau_a$ equals 0.038.
We thereby use this optimal setup as the default for audio pseudo label generation.
It is noteworthy that the event-level F-score is only around 64\% if simply assigning the video labels to all the audio segments (without prompt).
In contrast, this metric is around 72\% for our generated audio pseudo labels.
This reveals the vital role of segment-level event identification.

\begin{table}[t]\small
\caption{\textbf{Parameter study of the $K$ and scaling factor $\alpha$ used in the VISUAL pseudo label denoising.} 
Different values of $K$ and $\alpha$ are tested for the segment-wise visual pseudo label denoising.
The segment-level and event-level F-scores of the denoised visual pseudo labels are reported. 
The last column is the average result. \textit{`-' denotes the result of the visual pseudo label generated by PLG without label denoising.} This experiment is conducted on the validation set of the LLP dataset.} 
\centering
\label{tab:parameter_study_PLD}
\small
\resizebox{\linewidth}{!}{
 \begin{tabular}{p{1.2cm}<{\centering}p{1.2cm}<{\centering}p{1.2cm}<{\centering}p{1.2cm}<{\centering}p{1.2cm}<{\centering}}
\toprule[0.8pt]
\multicolumn{2}{c}{Parameter setup} & \multicolumn{1}{c}{\multirow{2}{*}{Segment. (V)}}  & \multicolumn{1}{c}{\multirow{2}{*}{
Event. (V)}}  & \multicolumn{1}{c}{\multirow{2}{*}{Average}}  \\ \cmidrule{1-2}
\multicolumn{1}{c}{$K$} & \multicolumn{1}{c}{$\alpha$} & & & \\ \midrule
- & - & 71.08 & 64.82 & 67.95 \\ \midrule
4 & \multirow{3}{*}{30} & 72.45 & 67.82 & 70.13 \\
\textbf{5} & & \textbf{72.99} & \textbf{68.28} & \textbf{70.63} \\
6 & & 72.17 & 66.90 & 69.53 \\ \midrule
\multirow{4}{*}{5} & {20} & {72.85} & {68.10} & {70.47} \\
& \textbf{30} & \textbf{72.99} & \textbf{68.28} & \textbf{70.63} \\
& 40 & 72.82 & 68.12 & 70.47 \\
\toprule[0.8pt]
\end{tabular}}
\end{table}

\textbf{Study of the $K$ and $\alpha$ in PLD.}
For each predicted event category, the top-$K$ smallest forward loss along the temporal dimension is magnified by $\alpha$ and used as the threshold to determine which segments' pseudo labels should be refined (Eq.~\ref{eq:PLD}).
\textit{The segment-level and event-level F-scores of the events are used to evaluate the quality of the denoised pseudo labels. 
}
For the visual pseudo label denoising, the results in Table~\ref{tab:parameter_study_PLD} indicate that denoised visual pseudo labels ensure significantly better results than the original labels generated by PLG.
In particular, the event-level F-score is improved by 3.46\%. 
Observing Table~\ref{tab:parameter_study_PLD}, the optimal setup are $K = $ 5 and $\alpha = $ 30.
Under this setup, the segment-level and event-level F-scores of the visual pseudo labels of the validation set achieve 72.99\% and 68.28\%, respectively. 
For the audio pseudo label denoising, as shown in Table~\ref{tab:parameter_study_APLD}, the denoised audio pseudo labels are slightly better than the pseudo labels generated by PLG under the optimal setup ($K = $ 6, $\alpha$ = 400).
As discussed in Sec.~\ref{sec:method_PLD}, PLD is proposed to alleviate the potentially discontinuous pseudo-event labels that happened during PLG and provide better temporal boundaries of the events. We argue that the discontinuity of pseudo labels of audio events rarely occurs due to the temporal characteristics of audio data, thus leading to a slight improvement for the audio modality as shown in Table~\ref{tab:parameter_study_APLD}. 
Besides, 
from Tables ~\ref{tab:parameter_study_PLD} and ~\ref{tab:parameter_study_APLD}, we observe an interesting phenomenon that the segment-level and event-level F-scores of audio pseudo labels without PLD (80.32\% and 71.54\%) remain superior to those of the denoised visual pseudo labels (72.99\% and 68.28\%).
This suggests the high quality of audio pseudo labels generated by PLG and underscores the greater difficulty in denoising visual pseudo labels. 
We ultimately strike a balance between the second computational costs and denoising improvements and refrain from applying PLD to the audio modality in our experiment setup.

\begin{table}[t]\small
\caption{\textbf{Parameter study of the $K$ and scaling factor $\alpha$ used in the AUDIO pseudo label denoising.} 
Different values of $K$ and $\alpha$ are tested for the segment-wise audio pseudo label denoising.
The segment-level and event-level F-scores of the denoised audio pseudo labels are reported. 
The last column is the average result. \textit{`-' denotes the result of the audio pseudo label generated by PLG without label denoising.} This experiment is conducted on the validation set of the LLP dataset.} 
\centering
\label{tab:parameter_study_APLD}
\small
\resizebox{\linewidth}{!}{
 \begin{tabular}{p{1.2cm}<{\centering}p{1.2cm}<{\centering}p{1.2cm}<{\centering}p{1.2cm}<{\centering}p{1.2cm}<{\centering}}
\toprule[0.8pt]
\multicolumn{2}{c}{Parameter setup} & \multicolumn{1}{c}{\multirow{2}{*}{Segment. (A)}}  & \multicolumn{1}{c}{\multirow{2}{*}{
Event. (A)}}  & \multicolumn{1}{c}{\multirow{2}{*}{Average}}  \\ \cmidrule{1-2}
\multicolumn{1}{c}{$K$} & \multicolumn{1}{c}{$\alpha$} & & & \\ \midrule
- & - & 80.32 & 71.54 & 75.93 \\ \midrule
5 & \multirow{3}{*}{400} & 79.63 & 70.88 & 75.25 \\
\textbf{6} & & \textbf{80.43} & \textbf{71.68} & \textbf{76.06} \\ 
7 & & 80.15 & 71.33 & 75.74 \\
\midrule
\multirow{3}{*}{6} & 300 & 80.16 & 71.27 & 75.72 \\
& \textbf{400} & \textbf{80.43} & \textbf{71.68} & \textbf{76.06} \\
& 500 & 80.40 & 71.27 & 75.72 \\
\toprule[0.8pt]
\end{tabular}}
\end{table}

\begin{figure*}[!htp]
\centering
\includegraphics[width=.98\textwidth]{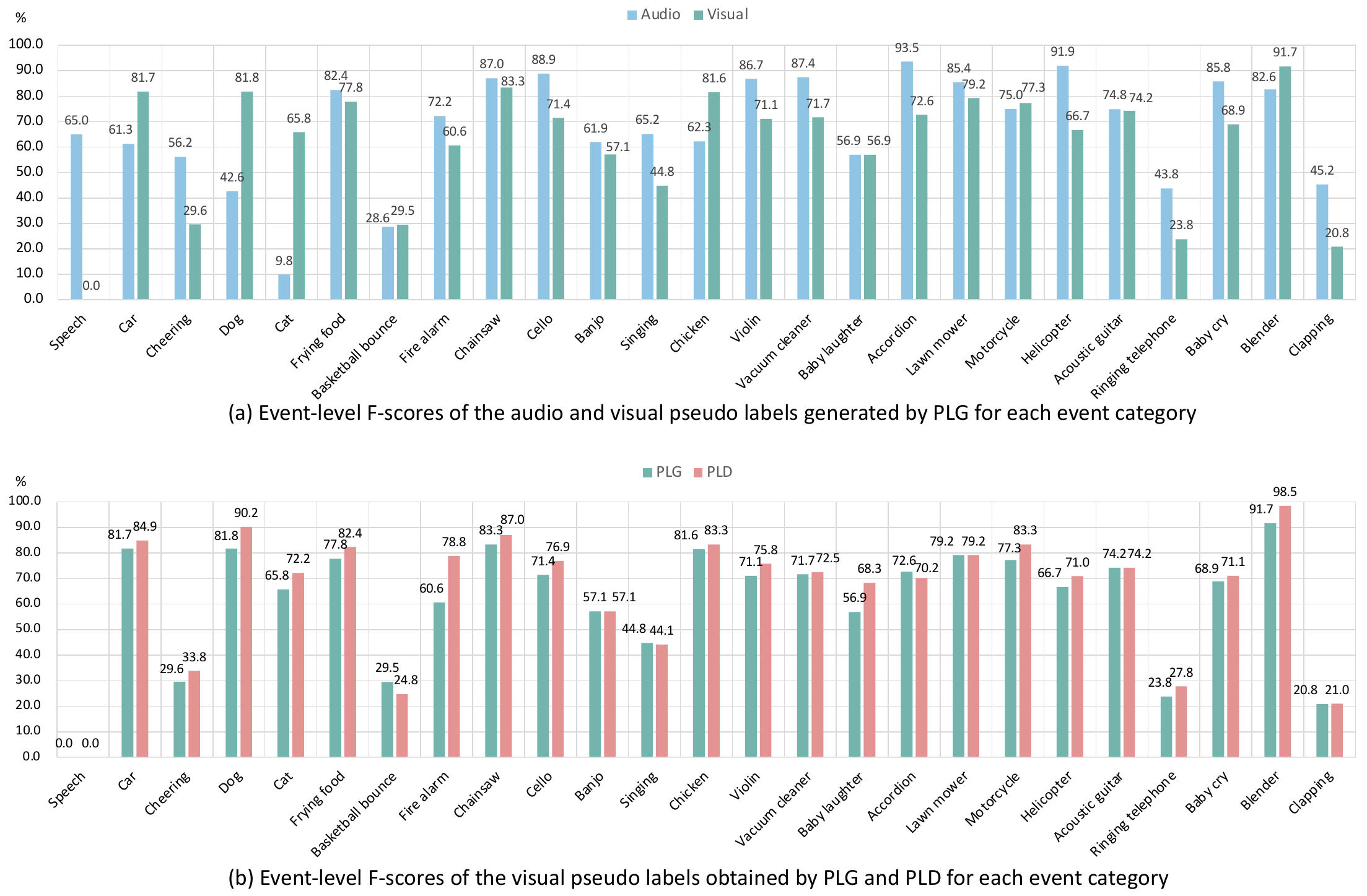}
\caption{\textbf{Event-level F-scores of pseudo labels for each event category.}
(a) We display the event-level F-scores of audio and visual pseudo labels generated by PLG.
(b) Compared to PLG, PLD further improves the event-level F-scores for most categories, providing more accurate visual pseudo labels.
All the results are reported on the validation set of the LLP dataset.
}
\vspace{3ex}
\label{fig:vis_event_score_of_each_class}
\end{figure*}

\begin{table*}[!htp]\small
\caption{\textbf{Ablation study of the main modules.}
Id-\ding{192} denotes the performance of the baseline backbone HAN~\citep{tian2020HAN}.
$\mathcal{L_S}$ is the proposed richness-aware loss (Eq.~\ref{eq:RL-segment_level}).
$\mathcal{L'_S}$ is a native loss that simply computes the binary cross entropy loss of the prediction and pseudo label.
We report the results on the test set of the LLP dataset.}
\centering
\label{tab:ablation}
\resizebox{\linewidth}{!}{
 \begin{tabular}{p{0.2cm}<{\centering}p{0.6cm}<{\centering}p{0.8cm}<{\centering}p{0.6cm}<{\centering}p{0.9cm}<{\centering}p{0.9cm}<{\centering}p{0.9cm}<{\centering}p{1.2cm} <{\centering}p{1.2cm} <{\centering}p{0.9cm}<{\centering}p{0.9cm}<{\centering}p{0.9cm}<{\centering}p{1.2cm}<{\centering}p{1.2cm}<{\centering}}
\toprule[0.8pt]
 \multirow{2}{*}{Id} & \multicolumn{3}{c}{Main modules} & \multicolumn{5}{c}{Segment-level} & \multicolumn{5}{c}{Event-level}  \\
 \cmidrule(r){2-4} \cmidrule(r){5-9} \cmidrule{10-14}
  & PLG & PLE & PLD & A & V & AV & Type@AV & Event@AV & A & V & AV & Type@AV & Event@AV \\
\midrule
\ding{192} & \ding{56} & \ding{56} &   \ding{56} & 60.1 & 52.9 & 48.9 & 54.0 & 55.4 & 51.3 & 48.9 & 43.0 & 47.7 & 48.0 \\
\midrule
 \ding{193} & \ding{52} & \ding{56} &   \ding{56} & 59.8 & 64.1 & 57.5 & 60.5 & 58.3 & 50.8 & 60.2 & 50.7 & 53.9 & 49.3 \\
\midrule
 \ding{194} & \ding{52}  & \ding{52}-$\mathcal{L'_S}$ & \ding{56} & 61.5 & {64.7} & 58.6 & 61.6 & 60.0 & 54.5 & 61.0 & 52.4 & 55.9 & 52.7 \\
\ding{195} & \ding{52}  & \ding{52}-$\mathcal{L_S}$ & \ding{56} & 61.2 & 65.8 & 59.1 & 62.0 & 60.2 & 54.8 & 62.4 & 52.6 & 56.6 & 53.3 \\
 \midrule
\ding{196} &  \ding{52}  & \ding{52} & \ding{52} & \textbf{62.4} & \textbf{66.7} & \textbf{60.3} & \textbf{63.1} & \textbf{61.4} & \textbf{55.7} & \textbf{63.3} & \textbf{53.7} & \textbf{57.6} & \textbf{54.3}  \\ 
\toprule[0.8pt]
\end{tabular}}
\end{table*}

\subsection{Ablation Studies}\label{exp:ablations}
In this section, we provide some ablation studies to explore the impact of each module in our method.
The experimental results are shown in Table~\ref{tab:ablation}.
The row with id-\ding{192} denotes the performance of the baseline HAN~\citep{tian2020HAN}.

\textbf{Impact of the PLG.} 
To further verify the benefits of PLG, we use the generated pseudo labels to supervise the model training.
Note that the vanilla HAN (id-\ding{192} in Table~\ref{tab:ablation}) is trained with the video-level pseudo label obtained by using label smoothing on the given weak label (Eq.~\ref{eq:HAN_loss}).
For a fair comparison, we only use the video-level pseudo labels generated by PLG as the model supervision (Eq.~\ref{eq:RL-video_level}).
 {As shown in row-\ding{193} of Table~\ref{tab:ablation}, utilizing the video-level pseudo label generated by our PLG significantly improves the visual event parsing performances.
The visual metric (V) increases from 52.9\% to 64.1\% at the segment level while from 48.9\% to 60.2\% at the event level.
These improvements reflect that our PLG generates more accurate video-level pseudo labels for the visual modality, better distinguishing the event categories and guiding the model training.
The improvement in audio event parsing is not pronounced in this situation.
We anticipate that the temporally continuous audio segments are more challenging to distinguish under weak video-level supervision.
Additionally, the visual features can encapsulate more distinct event semantics, thereby promoting model optimization that is more beneficial to the visual modality.
Even so, the utilization of more fine-grained, {segment-level} pseudo labels generated by our PLG (see ids \ding{174} and \ding{175} in Table ~\ref{tab:ablation}) significantly enhances both the audio and visual event parsing performances.
}


Our PLG is able to generate high-quality pseudo labels at the segment level, which can be verified by the results shown in Tables~\ref{tab:parameter_study_PLG} and ~\ref{tab:parameter_study_APLG}.
In Fig.~\ref{fig:vis_event_score_of_each_class} (a), we further display the event-level F-scores of the generated audio and visual pseudo labels of each event category and provide more discussions.
As seen, the audio and visual pseudo labels have satisfactory F-scores for most of the categories. The highest F-score is 93.5\% for audio event \textit{Accordion} and 91.7\% for visual event \textit{Blender}, respectively. 
Besides, we also find that each modality faces some intractable event categories, such as the \textit{speech} for visual modality and \textit{cat} for audio modality.
We argue this is caused by the unbalanced data distribution and some categories are particularly difficult for visual recognition, such as \textit{speech}, \textit{cheering}, and \textit{clapping}.
Nevertheless, our PLG generally provides reliable audio and visual pseudo labels from both the video level and segment level, ensuring better model learning.

\begin{table}[!hbp]\small
\caption{\textbf{Richness-aware loss $\mathcal{L_S}$ under different configurations.}
$\mathcal{SR}$ and $\mathcal{CR}$ denote that we only compute $\mathcal{L_S}$ with the segment richness and category richness alignment, respectively.
} 
\centering
\label{tab:rl_loss_ablation}
\small
\resizebox{\linewidth}{!}{
 \begin{tabular}{p{.5cm}<{\centering}p{.5cm}<{\centering}p{1.2cm}<{\centering}p{1.6cm}<{\centering}p{1.2cm}<{\centering}p{1.6cm}<{\centering}}
\toprule[0.8pt]
\multicolumn{2}{c}{Loss $\mathcal{L_S}$ } & \multicolumn{2}{c}{Segment-level} & \multicolumn{2}{c}{Event-level} \\
\cmidrule(r){1-2} \cmidrule(r){3-4} \cmidrule(r){5-6}
$\mathcal{SR}$ & $\mathcal{CR}$ & Type@AV & {Event@AV} & Type@AV & Event@AV \\ \midrule
\ding{56}& \ding{56} & 60.5 & 58.3 & 53.9 & 49.3  \\
\ding{56} & \ding{52} & 61.8 & \textbf{60.2} & 56.4 & 52.9 \\
\ding{52}& \ding{56} & 61.3 & 59.6 & 56.1 & 52.6 \\
\ding{52} & \ding{52} & \textbf{62.0} & \textbf{60.2} & \textbf{56.6} & \textbf{53.3} \\
\toprule[0.8pt]
\end{tabular}}
\end{table}

\textbf{Impact of the PLE.}
Our PLE uses the proposed richness-aware loss $\mathcal{L_S}$ in Eq.~\ref{eq:RL-segment_level} 
to exploit the pseudo labels from segment-level, which is taken as a complement to the video-level supervision. At first, we make an ablation study to explore the effect of the respective richness component. 
As shown in Table~\ref{tab:rl_loss_ablation}, ``$\mathcal{SR}$'' and ``$\mathcal{CR}$'' denote 
the segment richness loss and category richness loss between the predictions and pseudo labels, respectively.
From Table~\ref{tab:rl_loss_ablation}, we can find that each of them can effectively improve the model performance since the studied AVVP task requires distinguishing both the video segments and the event categories. 
When both types of richness information are used, the pseudo labels fully demonstrate the capability for model optimization.
To further validate its superiority, we compare it with a native variant that directly computes the binary cross entropy loss between the predictions and the pseudo labels, denoted as $\mathcal{L'_S} = \mathcal{L}_{\text{bce}}(\bm{P}^v, \hat{\bm{Y}}^v) + \mathcal{L}_{\text{bce}}(\bm{P}^a, \hat{\bm{Y}}^a)$.
As shown in the row-\ding{194} and \ding{195} of Table~\ref{tab:ablation}, both $\mathcal{L'_S}$ and the proposed $\mathcal{L_S}$ are beneficial for the audible video parsing since they all provide segment-level supervision.
Nevertheless, the proposed RL loss is more helpful. 
 {
The conventional cross-entropy loss relies on \textit{`hard' segment-wise} alignments between predictions and pseudo labels.
In contrast, our proposed richness-aware loss exploits the pseudo labels by aligning predictions from two \textit{independent} dimensions: category-richness ($cr$) and segment-richness ($sr$). 
According to the definitions of $cr$ (Eq.~\ref{eq:category_richness}) and $sr$ (Eq.~\ref{eq:segment_richness}), their values are expressed as percentages (\textit{`soft' ratios}) and are independent.
This design makes the model trained with our richness-aware loss automatically balance and utilize the soft supervisions from category-richness and segment-richness.
Experimental results shown in Table~\ref{tab:rl_loss_ablation} indicate the superiority of our flexible design of richness-aware loss.
}

\textbf{Impact of the PLD.}
The impact of PLD can be observed from two aspects.
On one hand, PLD provides more accurate pseudo labels than PLG. As the quality measurement of visual pseudo labels shown in Table~\ref{tab:parameter_study_PLD} on the validation set, the average F-score is 67.95\% for PLG while it is 70.63\% for PLD. 
In Fig.~\ref{fig:vis_event_score_of_each_class}(b), we show event-level F-scores for the visual pseudo labels obtained by PLG and PLD of each event category.
PLD further improves the F-scores for most categories (18/25), \eg, the metrics for events \textit{Fire alarm} and \textit{Blender} increase substantially by 18.2\% and 6.8\%, respectively.
On the other hand, visual pseudo labels generated by PLD are more helpful than PLG for model training. 
We update the visual pseudo labels as the new supervision to train the HAN model.
As shown in row-\ding{196} of Table~\ref{tab:ablation}, 
the model has superior performance on all types of event parsing. This again reveals that the visual pseudo labels obtained by PLD are more accurate than by PLG and can better supervise the multi-modal parsing model.
These results verify the effectiveness of the label denoising strategy in PLD.

\begin{table*}[t]\small
\caption{\textbf{Comparison with the state-of-the-arts.}
${\blacktriangle}$ represents these methods are all focused on generating better pseudo labels for the AVVP task and are all developed on the baseline HAN~\citep{tian2020HAN} backbone.
$\bigstar$ denotes we further implement our method with the more advanced visual and audio features extracted by CLIP and CLAP, respectively.
Results are reported on the test set of the LLP dataset.}
\centering
\label{tab:sota_comparison}
\resizebox{\linewidth}{!}{
 \begin{tabular}{p{4.4cm}<{\centering} p{0.8cm}<{\centering}p{0.8cm}<{\centering}p{0.8cm}<{\centering}p{1.2cm} <{\centering}p{1.2cm} <{\centering}p{1.cm}<{\centering}p{0.8cm}<{\centering}p{0.8cm}<{\centering}p{1.2cm}<{\centering}p{1.2cm}<{\centering}}
\toprule[0.8pt]
\multirow{2}{*}{Method} & \multicolumn{5}{c}{Segment-level} & \multicolumn{5}{c}{Event-level}  \\
 \cmidrule(r){2-6} \cmidrule(r){7-11}
 & A & V & AV & Type@AV & Event@AV & A & V & AV & Type@AV & Event@AV \\
\midrule
AVE~\citep{tian2018audio} & 47.2 & 37.1 & 35.4 & 39.9 & 41.6 & 40.4 & 34.7 & 31.6 & 35.5 & 36.5 \\
AVSDN~\citep{lin2019dual} & 47.8 & 52.0 & 37.1 & 45.7 & 50.8 & 34.1 & 46.3 & 26.5 & 35.6 & 37.7 \\
\rowcolor{gray!20}HAN~\citep{tian2020HAN} & 60.1 & 52.9 & 48.9 & 54.0 & 55.4 & 51.3 & 48.9 & 43.0 & 47.7 & 48.0 \\
MM-Pyramid~\citep{yu2022MMPyramid} & 60.9 & 54.4 & 50.0 & 55.1 & 57.6 & 52.7 & 51.8 & 44.4 & 49.9 & 50.5 \\
MGN~\citep{mo2022MGN} & 60.8 & 55.4 & 50.4 & 55.5 & 57.2 & 51.1 & 52.4 & 44.4 & 49.3 & 49.1 \\
CVCMS~\citep{lin2021CVCMS} & 59.2 & 59.9 & 53.4 & 57.5 & 58.1 & 51.3 & 55.5 & 46.2 & 51.0 & 49.7 \\
DHHN~\citep{jiang2022DHHN} & 61.3 & 58.3 & 52.9 &  57.5 & 58.1 & 54.0 & 55.1 & 47.3 & 51.5 & 51.5 \\
\midrule
$^{\blacktriangle}${MA}~\citep{wu2021MA} & 60.3 & 60.0 & 55.1 & 58.9 & 57.9 & 53.6 & 56.4 & 49.0 & 53.0 & 50.6 \\
$^{\blacktriangle}$JoMoLD~\citep{cheng2022JoMOLD} & {61.3} & 63.8 & 57.2 & 60.8 & 59.9 & {53.9} & 59.9 & 49.6 & 54.5 & {52.5} \\ 
$^{\blacktriangle}$\textbf{VAPLAN (ours)} & \textbf{62.4} & \textbf{66.7} & \textbf{60.3} & \textbf{63.1} & \textbf{61.4} & \textbf{55.7} & \textbf{63.3} & \textbf{53.7} & \textbf{57.6} & \textbf{54.3}  \\
\hdashline
$^{\bigstar}$\textbf{VAPLAN (ours)} & \textbf{69.0} & \textbf{70.2} & \textbf{63.5} & \textbf{67.6} & \textbf{67.9} & \textbf{61.9} & \textbf{66.4} & \textbf{56.9} & \textbf{61.7} & \textbf{60.1}  \\ 
\toprule[0.8pt]
\end{tabular}}
\end{table*}

\begin{table*}[!t]\small
\caption{\textbf{Generalization of our method on other audio-visual video parsing backbones.}
Our method can generate reliable segment-level audio and visual pseudo labels which can be directly used for other methods in the AVVP task too.
We evaluate two representative backbones, namely the MGN~\citep{mo2022MGN} and MM-Pyramid~\citep{yu2022MMPyramid}.
The pseudo labels generated by our PLG and refined by our PLD consistently boost these models. Both PLG and PLD are also superior to the existing method MA~\citep{wu2021MA} that provides video-level pseudo labels.
The best and second-best results of each evaluation metric are \textbf{bold} and \underline{underlined}, respectively.
} 
\centering
\label{tab:PLG_PLD_on_other_methods}
\resizebox{\linewidth}{!}{
 \begin{tabular}{p{4.2cm}<{\centering} p{0.8cm}<{\centering}p{0.8cm}<{\centering}p{0.8cm}<{\centering}p{1.2cm} <{\centering}p{1.2cm} <{\centering}p{1.cm}<{\centering}p{0.8cm}<{\centering}p{0.8cm}<{\centering}p{1.2cm}<{\centering}p{1.2cm}<{\centering}}
\toprule[0.8pt]
\multirow{2}{*}{Method} & \multicolumn{5}{c}{Segment-level} & \multicolumn{5}{c}{Event-level}  \\
 \cmidrule(r){2-6} \cmidrule(r){7-11}
 & A & V & AV & Type@AV & Event@AV & A & V & AV & Type@AV & Event@AV \\
\midrule
\rowcolor{gray!20}MGN~\citep{mo2022MGN} & \underline{60.8} & 55.4 & 50.4 & 55.5 & 57.2 & \textbf{51.1} & 52.4 & 44.4 & 49.3 & 49.1 \\
MGN + MA & 60.2 & {61.9} & 55.5 & 59.2 & 58.7 & \underline{50.9} & {59.7} & 49.6 & 53.4 & \underline{49.9} \\
MGN + \textbf{PLG} & 60.1 & \underline{63.3} & \underline{56.5} & \underline{60.0} & \underline{58.9} & 50.3 & \underline{60.9} & \underline{50.2} & \underline{53.8} & 49.4 \\
MGN + \textbf{PLD} & \textbf{61.0} & \textbf{64.3} & \textbf{57.1} & \textbf{60.8} & \textbf{60.1} & \textbf{51.1} & \textbf{61.9} & \textbf{50.6} & \textbf{54.5} & \textbf{50.4} \\
\midrule
\rowcolor{gray!20}MM-Pyramid~\citep{yu2022MMPyramid} & {60.9} & {54.4} & 50.0 & 55.1 & 57.6 & {52.7}& 51.8 & 44.4 & 49.9 & 50.5 \\
MM-Pyramid + MA & \textbf{61.1} & {60.3} & 55.8 & 59.7 & 59.1 & 53.8 & {56.7} & \underline{49.4} & 54.1 & {51.2} \\
MM-Pyramid + \textbf{PLG} & 60.2 & \underline{65.4} & \underline{58.3} & \underline{61.3} & \underline{60.1} & \underline{54.5} & \underline{62.0} & \textbf{52.8} & \underline{56.4} & \underline{53.0} \\
MM-Pyramid + \textbf{PLD} & \underline{61.0} & \textbf{66.4} & \textbf{58.5} & \textbf{62.0} & \textbf{60.9} & \textbf{55.0} & \textbf{63.0} & \textbf{52.8} & \textbf{56.9} & \textbf{53.4} \\
\toprule[0.8pt]
\end{tabular}}
\end{table*}

\subsection{Comparison with the State-of-the-arts}
We 
report the performance of our VAPLAN on the test set of the LLP dataset. The comparison results with existing methods are shown in Table~\ref{tab:sota_comparison}.
Our method achieves superior performance on all types of event parsing.
\textbf{First}, compared to the baseline HAN~\citep{tian2020HAN} on which our method is developed, our method significantly improves the performance.
Especially for the visual event parsing (\textbf{V} in the table), the segment-level metric is lifted from 52.9\% to 66.7\% ($\uparrow$ 13.8\%), and the event-level metric is improved from 48.9\% to 63.3\% ($\uparrow$ 14.4\%).
\textbf{Second}, our method outperforms other competitors on the track of generating pseudo labels for the AVVP task.
As shown in the low part of Table~\ref{tab:sota_comparison}, our method generally exceeds the previous state-of-the-art JoMoLD~\citep{cheng2022JoMOLD} by about 1.5 points for the audio event parsing, and around 3 points for the visual event and audio-visual event parsing.
Both JoMoLD~\citep{cheng2022JoMOLD} and MA~\citep{wu2021MA} generate audio-visual pseudo labels from the video level, while our method can provide audio-visual pseudo labels from a more fine-grained segment level.
Our video parsing model can be better supervised and optimized, resulting in better performance.
\textbf{Furthermore}, we report the result of our method
using the visual and audio features respectively extracted by CLIP and CLAP.
As shown in the last row of Table~\ref{tab:sota_comparison}, all types of event parsing performance can be further significantly improved.
In particular, the audio event parsing benefits more from such advanced feature representations.
As shown, its performance improves by 6.6\% and 6.2\% for the segment-level and event-level F-scores, respectively.
These improvements demonstrate the effectiveness and superiority of our method.

\begin{figure*}[t]
\centering
\includegraphics[width=\textwidth]{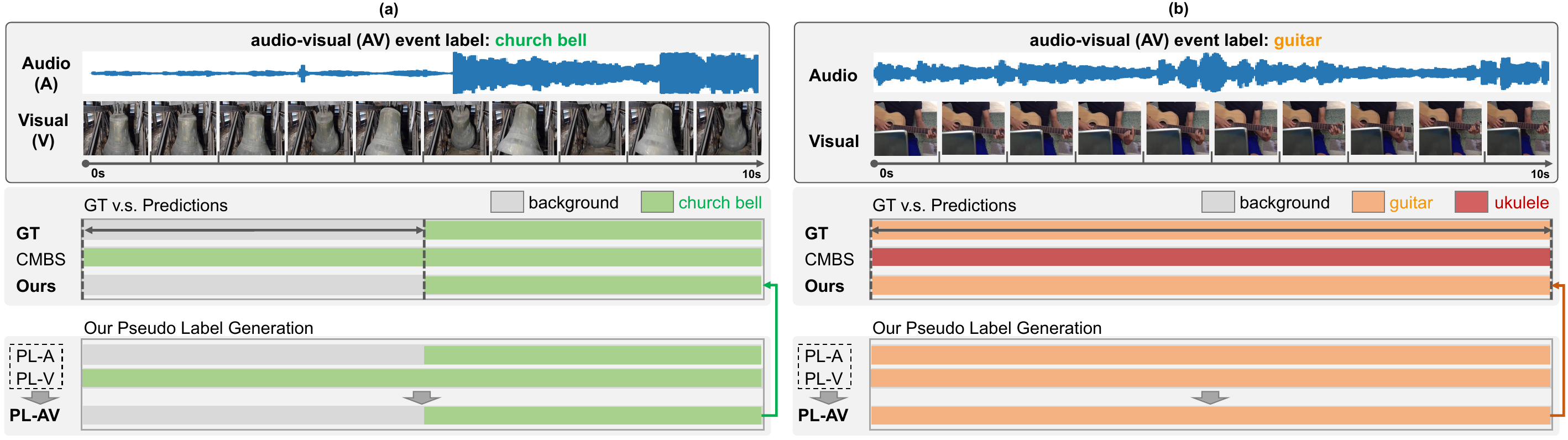}
\caption{ {\textbf{Qualitative examples for the weakly-supervised audio-visual event localization task.}
This task aims to temporally locate those segments containing events that are both audible and visible.
The previous state-of-the-art method, CMBS~\citep{xia2022cross}, utilizes only the video-level weak labels for model training and predictions.
In contrast, our method can generate high-quality segment-level pseudo labels, offering fine-grained supervision during training and producing more accurate localization results. 
``GT'' denotes the ground truth.
``PL-A'' and ``PL-V'' represent our segment-level pseudo labels for the audio and visual modalities, respectively.
The audio-visual event pseudo labels (``PL-AV'') result from the intersection of ``PL-A'' and ``PL-V''.
Our method surpasses the vanilla CMBS model in distinguishing between the background and audio-visual events (a) as well as among different audio-visual event categories (b).}}
\label{fig:vis_ave_results}
\end{figure*}

\subsection{Generalization of Our Method}
\noindent\textbf{Generalization on other AVVP backbones.} A core contribution of our method is that it can provide high-quality segment-level audio and visual pseudo labels, which then better guide the model optimization.
Our method can also be applied to other existing backbones in the AVVP task. To explore its impact, we examine two recently proposed networks, \ie, MGN~\citep{mo2022MGN} and MM-Pyramid~\citep{yu2022MMPyramid}.
Specifically, we train the models using the pseudo labels generated by our PLG and refined by our PLD, respectively.
The experimental results are shown in Table~\ref{tab:PLG_PLD_on_other_methods}.
Both PLG and PLD significantly boost the vanilla models, especially in the visual event and audio-visual event parsing. Take the MM-Pyramid~\citep{yu2022MMPyramid} method for example, the segment-level visual event parsing performance is improved from 54.4\% to 65.4\% and 66.4\% by using our PLG and PLD, respectively.
PLD is superior due to the additional label denoising strategy.
Such improvements can also be observed for MGN~\citep{mo2022MGN}.
Besides, it is worth noting that these two backbones perform better when combined with our (segment-level) pseudo labels than the (video-level) pseudo labels generated by the previous method MA~\citep{wu2021MA}. 
These results again indicate that our method is able to provide better fine-grained pseudo labels and demonstrate the superiority and generalizability of our method.

\begin{table}[t]\small
\caption{\textbf{Generalization of our method on the weakly-supervised audio-visual event localization task.}
Given the only video-level event label, this task needs to localize the temporal video segments that contain the audio-visual event, \ie, the audio and visual segments simultaneously describe the same event. We extend our pseudo label generation strategy to this task and generate segment-level event labels. We test several SOTA models on this task, namely AVEL~\citep{tian2018audio}, PSP~\citep{zhou2021psp}, and CMBS~\citep{xia2022cross}. All of them can be further improved using our segment-level pseudo labels as the objective.
This experiment is conducted on the AVE~\citep{tian2018audio} dataset.} 
\centering
\label{tab:avel_results}
\small
 \begin{tabular}{lp{2.5cm}<{\centering}p{2.8cm}<{\centering}}
\toprule[0.8pt]
\multicolumn{1}{l}{\multirow{2}{*}{Method}} & \multicolumn{2}{c}{label objective}  \\ \cmidrule{2-3}
& \multicolumn{1}{c}{video-level} & \multicolumn{1}{c}{segment-level (\textbf{ours})} \\ \midrule
AVEL & 67.1 & \textbf{69.2}$_{(+2.1)}$  \\
PSP& 72.1 & \textbf{74.3}$_{(+2.2)}$ \\
CMBS & 72.2 & \textbf{74.4}$_{(+2.2)}$ \\
\toprule[0.8pt]
\end{tabular}
\end{table}

\noindent\textbf{Generalization on the AVEL task.}
We also extend our pseudo label generation strategy to a related audio-visual event localization (AVEL) task.
We explore the challenging weakly-supervised setting where the model needs to localize those video segments containing the audio-visual events (an event is both audible and visible) given only the video-level event category label.
Previous AVEL methods merely use the known video-level labels as the objective for model training.
Here we try to generate segment-level pseudo labels for this task as we did for the weakly-supervised AVVP task.
Similarly, we use the pretrained CLIP and CLAP models to generate segment-level visual and audio pseudo labels, respectively.
The audio-visual event pseudo labels are the intersection of them.
In this way, we know if there is an audio-visual event in each video segment.
Then such segment-level pseudo labels can be used as a new objective to supervise the model training.
We test three representative audio-visual event localization methods whose official codes are available, namely the AVEL~\citep{tian2018audio}, PSP~\citep{zhou2021psp} and CMBS~\citep{xia2022cross}. 
We conduct experiments on the corresponding AVE~\citep{tian2018audio} dataset and the results are shown in Table~\ref{tab:avel_results}.
The second column shows the performance of vanilla models with only the video-level supervision.
The last column shows that these models can be significantly improved by around 2 points when using our segment-level pseudo labels.

\begin{figure*}[!htp]
\centering
\includegraphics[width=\textwidth]{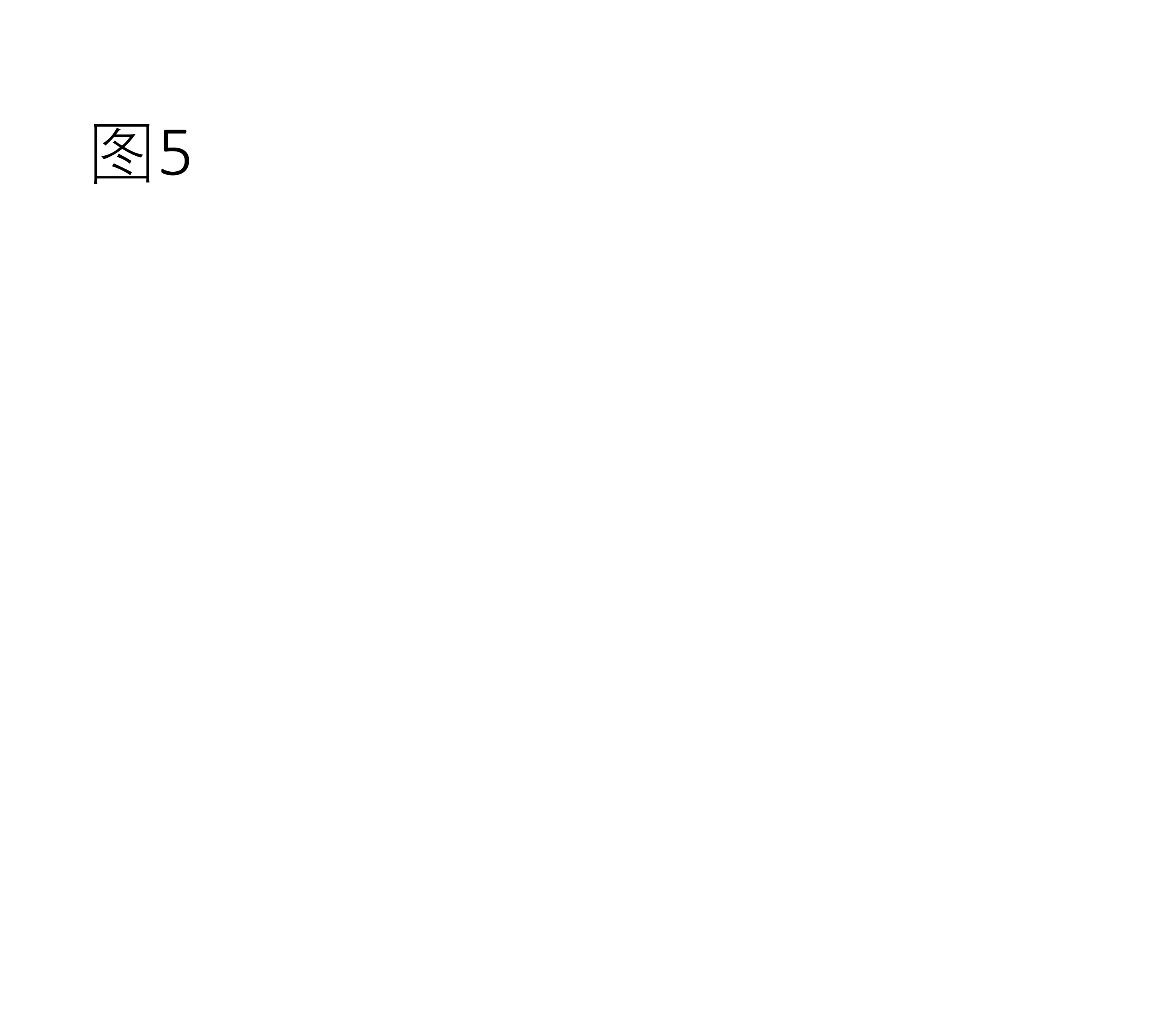}
\caption{\textbf{Qualitative examples of the audio-visual video parsing using different methods.}
We compare our method with the HAN~\citep{tian2020HAN}, MA~\citep{wu2021MA} and JoMoLD~\citep{cheng2022JoMOLD}.
``GT'' denotes the ground truth.
Our method successfully recognizes that there is only one visual event \textit{violin} in (a) or \textit{basketball bounce} in (b). Our method is also more accurate in parsing the audio events and audio-visual events, providing better temporal boundaries of the events.}
\label{fig:vis_parsing_results}
\end{figure*}

\begin{figure*}[!t]
\centering
\includegraphics[width=0.8\textwidth]{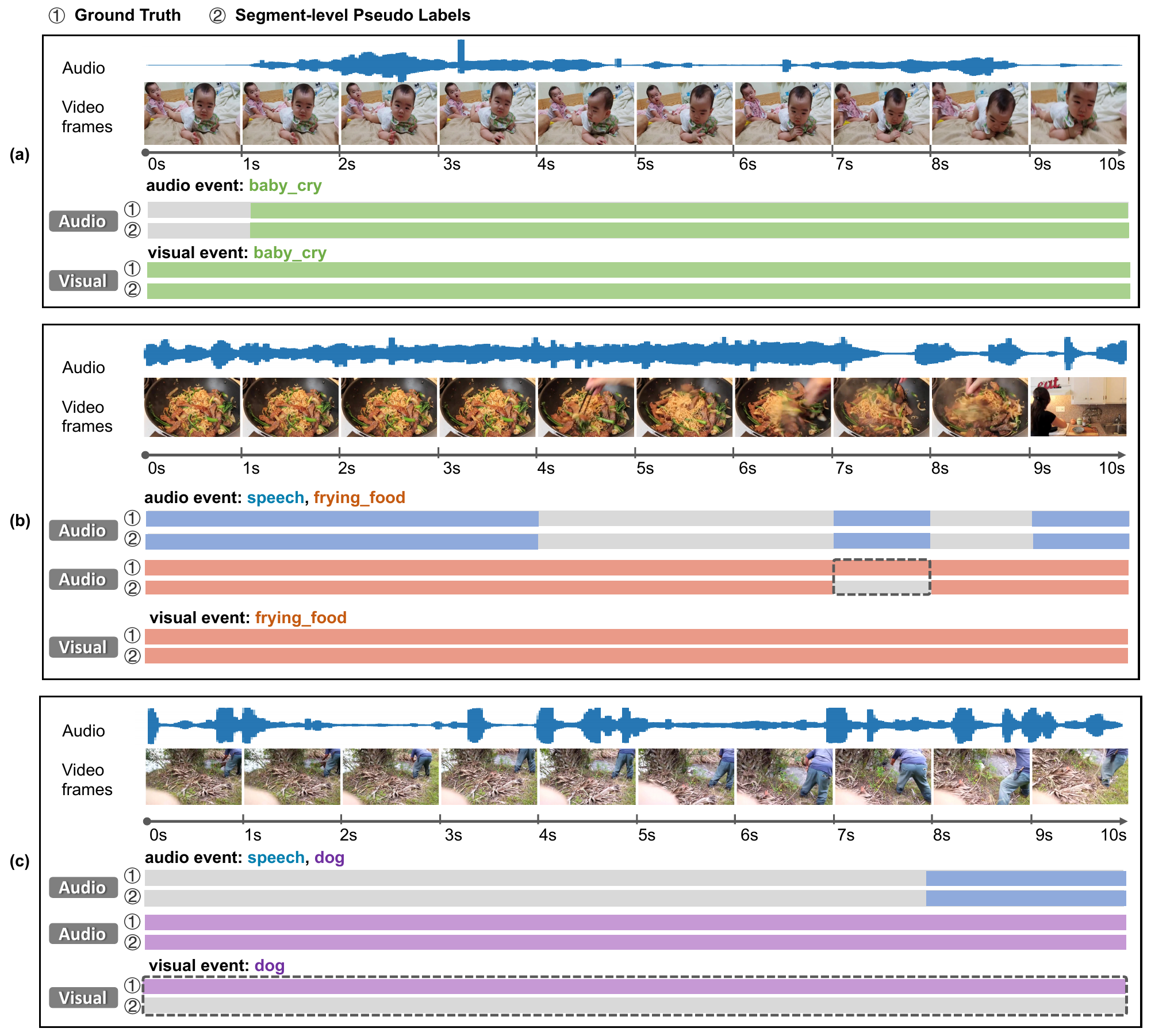}
\caption{\textbf{Typical and challenging visualization examples of the generated audio and visual pseudo labels.}
``\ding{192}'' and ``\ding{193}'' denote the ground truth and the obtained pseudo labels, respectively.
(a) In these typical cases where the events are clearly represented in the audio and visual signals, our method can generate accurate segment-level pseudo labels.
We also display some challenging examples: the audio event is mixed with other sounds (b) or the visual event is hard to perceive (c).
In general, our method can provide satisfactory audio and visual pseudo labels.
}
\label{fig:vis_gt_pseudo_labels}
\end{figure*}

\begin{figure*}[!t]
\centering
\includegraphics[width=0.8\textwidth]{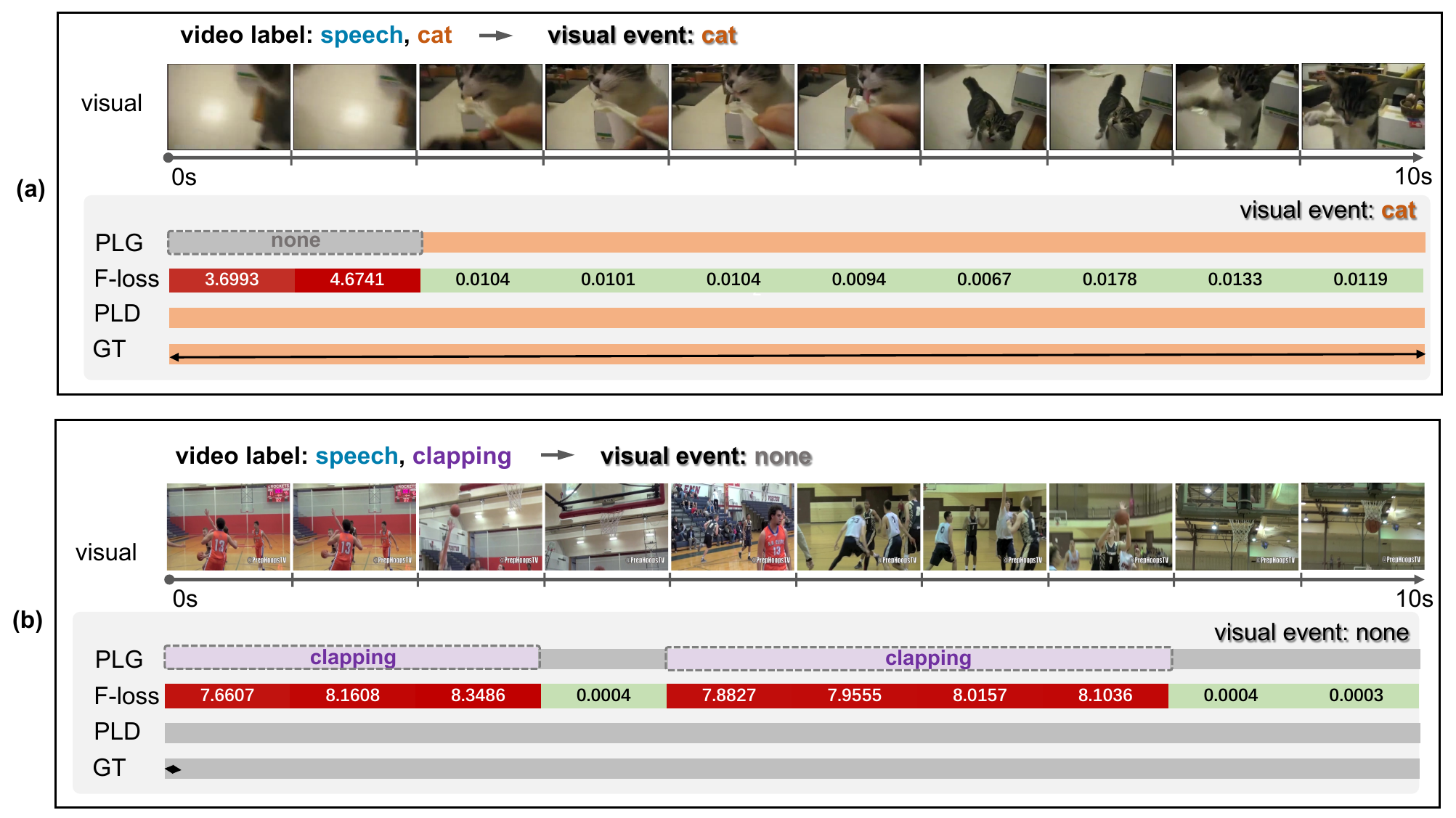}
\caption{\textbf{Qualitative visualization examples of the pseudo label denoising.}
Here, we take the visual modality as an example since it faces more challenges in both pseudo label generation and denoising processes.
``GT'' denotes the ground truth.
``F-loss'' represents the forward loss between the model predictions and the pseudo labels generated by PLG (Eq.~\ref{eq:fw_loss}).
PLG basically disentangles the visual event(s) from the weak video label, yielding well-defined segment-wise event categories.
Additionally, PLD helps alleviate potential label noise for those segments along the timeline in the same video whose pseudo labels generated by PLG suffer abnormally large loss values.
The improved labels are highlighted by the dotted box. 
}
\label{fig:vis_pseudo_labels}
\end{figure*}

 {We also present some qualitative examples for a more intuitive comparison. As shown in Fig.~\ref{fig:vis_ave_results} (a), the audio-visual event \textit{church bell} occurs exclusively in the last five video segments. The previous state-of-the-art method, CMBS, incorrectly assumes this event to be present in the first five segments as well. In contrast, our method yields accurate localization results.
The reason is that vanilla CMBS relies solely on the known weak event label (video-level) to supervise model training, while our method is capable of generating high-quality pseudo labels at the segment level. In the lower part of Fig.~\ref{fig:vis_ave_results} (a), we illustrate our pseudo label generation process. Our method accurately identifies that the \textit{church bell} event exists in all the visual segments but is present only in the last five audio segments, which results in the precise audio-visual event pseudo label and then better supervises the model training and predictions. 
Similar benefits can also be observed from Fig.~\ref{fig:vis_ave_results} (b), the vanilla CMBS incorrectly classifies the audio-visual event \textit{guitar} to be the \textit{ukulele}. In contrast, our method can generate accurate segment-level pseudo labels, thereby ensuring superior predictions.
These results again verify the generalization of our method and we believe our method can also help to address other related audio-visual tasks lacking fine-grained supervision.}

\subsection{Qualitative Results on the  AVVP task}\label{exp:qualitative}

\textbf{Visualization examples of the audio-visual video parsing.} We first display some qualitative video parsing examples in Fig.~\ref{fig:vis_parsing_results}.
We compare our method with HAN~\citep{tian2020HAN}, MA~\citep{wu2021MA}, and JoMoLD~\citep{cheng2022JoMOLD}.
Both MA and JoMoLD are developed on the HAN and try to generate video-level pseudo labels for better model training.
As shown in Fig.~\ref{fig:vis_parsing_results} (a), two events exist in the video, \ie, \textit{speech} and \textit{violin}, while the visual event only contains the \textit{violin}.
For audio event parsing, although all methods correctly recognize the two events occurring in the audio track, our method locates more exact temporal segments.
Also, our method accurately recognizes the visual event \textit{violin} and provides superior audio-visual event parsing. In Fig.~\ref{fig:vis_parsing_results} (b), both the events \textit{speech} and \textit{basketball bounce} exist in the video. 
All methods miss the audio event \textit{speech}.
The reason may be that the \textit{speech} event only happens in the second segment and the audio signal contains some noise from outdoors. It is hard to distinguish them. For visual and audio-visual event parsing, only our method provides satisfactory prediction for the audio event \textit{basketball bounce}. 
Although our method incorrectly identifies that the third segment contains this event, we argue that there may be an annotation mistake. 
The basketball player in this segment is clearer than in the second segment. If true, our result is more correct.
These video samples demonstrate the superiority of our method, which leverages high-quality segment-level pseudo labels to better supervise model training.

\textbf{Visualization examples of the obtained pseudo labels.} 
In this part, we display the pseudo labels of some typical and challenging video samples. Our method is able to provide high-quality segment-level audio and visual pseudo labels. 
As shown in Fig.~\ref{fig:vis_gt_pseudo_labels} (a), the \textit{baby cry} event is clearly represented in the video and our method successfully recognizes it in both audio and visual tracks. The temporal boundaries of the generated pseudo labels highly match the ground truth. Our method performs well in handling similar cases with explicit audio and visual event signals. Turning to Fig.~\ref{fig:vis_gt_pseudo_labels} (b), our method generates accurate pseudo labels for the visual event \textit{frying food} and audio event \textit{speech}.
The audio event \textit{frying food} in the eighth segment is not identified. The difficulty is that the sound of \textit{frying food} is mixed with the louder sound of \textit{speech}, which causes the \textit{frying food} event to be missed. The compound audio classification is still a challenging task in the community. 
In Fig.~\ref{fig:vis_gt_pseudo_labels} (c), our method satisfactorily generates segment-level pseudo labels for all the audio events but fails to recognize the visual event \textit{dog}.
The \textit{dog} in the visual frames is too small (located around the man's feet in the figure) to be identified. 
This situation is hard to judge even for a human annotator. The pseudo labels can be further explored in the future if considering more specific techniques for these challenging cases. Nevertheless, our method can generally provide reliable segment-level pseudo labels.

{\textbf{Visualization of the pseudo label denoising.}} 
As shown in Fig.~\ref{fig:vis_pseudo_labels}, we show two visualization examples to reflect the impact of pseudo label denoising. 
Here, we take the more challenging visual pseudo label denoising as an example.
As shown in Fig.~\ref{fig:vis_pseudo_labels} (a), the video-level label contains the events of \textit{speech} and \textit{cat}, where \textit{speech} 
does not exist in the visual modality. PLG successfully recognizes that only \textit{cat} event happens in the visual track. 
However, since the object is too blurry in the first two segments, the event \textit{cat} is incorrectly recognized. 
As a result, the forward loss values for these two segments are significantly greater, possibly 300 to 400 times larger than the other segments, as shown in the Fig.~\ref{fig:vis_pseudo_labels} (a). 
Contributing to the proposed label denoising (PLD) strategy, we make the correction.
Observing Fig.~\ref{fig:vis_pseudo_labels} (b), there are no visual events.
PLG mistakenly classifies a few segments as the event \textit{clapping} because the player's movements are complex in these segments. 
This inaccuracy is once again evident through the abnormally high forward losses.
PLD also rectifies these erroneous pseudo labels.
By analysis, the pseudo labels generated by PLG rely on the prior knowledge of event categories from the pretrained CLIP, while PLD benefits from an additional revision process (
-- the joint exploration of the predictions and pseudo labels through the forward loss calculation in each video) to possibly correct inaccurate segment-level pseudo labels in PLG.

\section{Conclusion}
We propose a Visual-Audio Pseudo LAbel exploratioN (VAPLAN) method for the weakly-supervised audio-visual video parsing task. 
VAPLAN is a new attempt to generate segment-level pseudo labels in this field, which starts with a pseudo label generation module that uses the reliable CLIP and CLAP models to determine the visual events and audio events occurring in each modality (at the segment level) as pseudo labels.
We then exploit the category richness and segment richness contained in the pseudo labels and propose a new richness-aware loss as fine-grained supervision for the AVVP task.
Furthermore, we propose a pseudo label denoising strategy to refine the visual pseudo labels and better guide the predictions. 
Qualitative and quantitative experimental results on the LLP dataset corroborate that our method can effectively generate and exploit high-quality segment-level pseudo labels. All these proposed techniques can be directly used in the community.
We also extend our method to a related weakly-supervised audio-visual event localization task and the experimental results verify the effectiveness and generalization of our method.
We believe this work will not only facilitate future research on the studied audio-visual video parsing task but also inspire other related audio-visual topics seeking better supervision.

\vspace{3mm}
\noindent\textbf{Data availability} \noindent The LLP dataset for the studied audio-visual video parsing is publicly available from the official website {\url{https://github.com/YapengTian/AVVP-ECCV20}}.
The AVE dataset for the audio-visual event localization task can be accessed at 
{\url{https://github.com/YapengTian/AVE-ECCV18}}.
Tables 1-9 and figures 3-7 were generated with our source codes, which will be released at our GitHub repository {\url{https://github.com/jasongief/VPLAN}}.

\begin{acknowledgements}
We would like to thank Dr. Liang Zheng for his constructive suggestions.
We also sincerely appreciate the anonymous reviewers for their positive feedback and professional comments.
\end{acknowledgements}



\bibliographystyle{spbasic}      
\bibliography{scob}   

\balance
\end{document}